\documentclass[letterpaper]{article} 
\usepackage{aaai2026}  
\usepackage{times}  
\usepackage{helvet}  
\usepackage{courier}  
\usepackage{graphicx} 
\usepackage[hyphens]{url}  
\usepackage{natbib}  
\usepackage{caption} 
\usepackage{algorithm}
\usepackage{algorithmic}
\usepackage{booktabs}
\usepackage{multirow}
\usepackage{amsmath}
\usepackage{amssymb}
\usepackage{hyperref}
\title{MCMoE: Completing Missing Modalities with Mixture of Experts for\\Incomplete Multimodal Action Quality Assessment}
\author{
    Huangbiao Xu\textsuperscript{\rm 1,\rm 2},
    Huanqi Wu\textsuperscript{\rm 1,\rm 2},
    Xiao Ke\textsuperscript{\rm 1,\rm 2}\thanks{Corresponding author.},
    Junyi Wu\textsuperscript{\rm 1,\rm 2},
    Rui Xu\textsuperscript{\rm 1,\rm 2},
    Jinglin Xu\textsuperscript{\rm 3}
}
\affiliations{
    \textsuperscript{\rm 1}Fujian Provincial Key Laboratory of Networking Computing and Intelligent Information Processing, College of Computer and Data Science, Fuzhou University, Fuzhou 350116, China\\
    \textsuperscript{\rm 2}Engineering Research Center of Big Data Intelligence, Ministry of Education, Fuzhou 350116, China\\
    \textsuperscript{\rm 3}School of Intelligence Science and Technology, University of Science and Technology Beijing, Beijing 100083, China\\
    kex@fzu.edu.cn,\{huangbiaoxu.chn, wuhuanqi135, xurui.ryan.chn, xujinglinlove\}@gmail.com, junyi.wu-1@outlook.com
}

\begin{document}

\maketitle

\begin{abstract}
    Multimodal Action Quality Assessment (AQA) has recently emerged as a promising paradigm. By leveraging complementary information across shared contextual cues, it enhances the discriminative evaluation of subtle intra-class variations in highly similar action sequences. However, partial modalities are frequently unavailable at the inference stage in reality. The absence of any modality often renders existing multimodal models inoperable. Furthermore, it triggers catastrophic performance degradation due to interruptions in cross-modal interactions. To address this issue, we propose a novel Missing Completion Framework with Mixture of Experts (MCMoE) that unifies unimodal and joint representation learning in single-stage training. Specifically, we propose an adaptive gated modality generator that dynamically fuses available information to reconstruct missing modalities. We then design modality experts to learn unimodal knowledge and dynamically mix the knowledge of all experts to extract cross-modal joint representations. With a mixture of experts, missing modalities are further refined and complemented. Finally, in the training phase, we mine the complete multimodal features and unimodal expert knowledge to guide modality generation and generation-based joint representation extraction. Extensive experiments demonstrate that our MCMoE achieves state-of-the-art results in both complete and incomplete multimodal learning on three public AQA benchmarks.
    \end{abstract}
    
    \begin{links}
        \link{Code}{https://github.com/XuHuangbiao/MCMoE}
    \end{links}
    
    \section{Introduction}
    Action quality assessment (AQA) has gained attention for its objective evaluation of action execution proficiency, with wide applications in sports, rehabilitation \cite{ding2023sedskill,bruce2024egcn++}, and skill determination \cite{xu2025dancefix}. Multimodal AQA \cite{xu2024vision,xu2025quality,tip/ZengZ24} has emerged as a promising paradigm beyond skeleton-based \cite{11,bruce2024egcn++} and vision-based \cite{ke2024two,xu2024fineparser,xu2024procedure,pami/XuYP25} methods. By leveraging complementary information from temporally aligned modalities, it better discriminates subtle intra-class variations in highly similar actions through enriched contextual cues.

    \begin{figure}[t]
        \centering
        \includegraphics[width=0.95\linewidth]{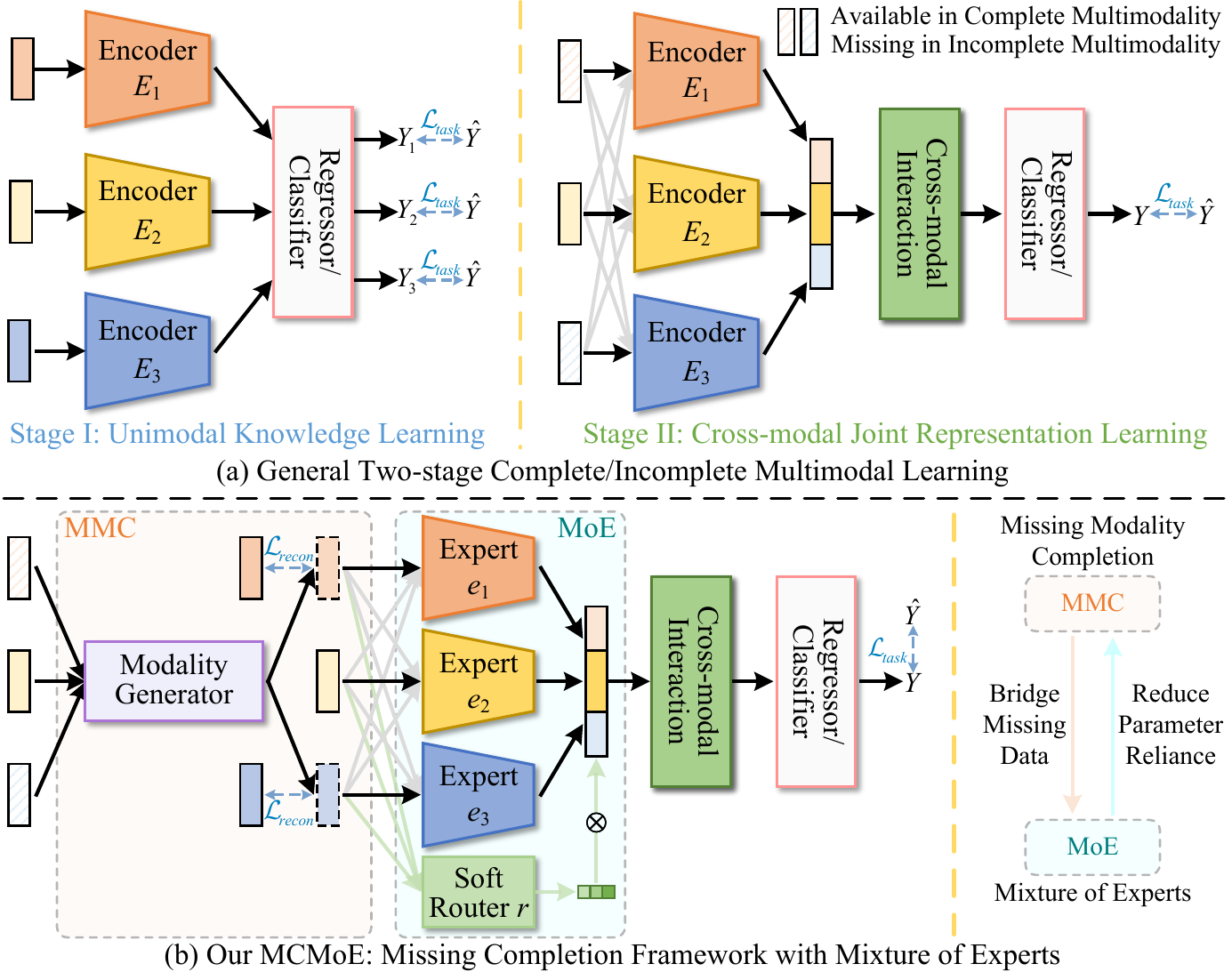}
        \caption{(a) Existing two-stage methods first learn unimodal features from complete multimodal data and then model cross-modal representations to address missing data, leading to higher training cost and complexity. (b) Our MCMoE unifies unimodal and joint representation learning in a single stage by exploiting the complementarity between modality completion and mixture of experts.}
        \label{fig:1}
    \end{figure}
    However, existing multimodal models often assume full modality availability during training and inference. Yet, real-world inference faces inevitable missing modalities due to sensor failures \cite{liu2021face}, environmental constraints \cite{12}, or privacy concerns \cite{jaiswal2020privacy}. Prior works \cite{wei2023mmanet,park2023cross} show that such incompleteness severely degrades performance by disrupting cross-modal interactions. Moreover, diverse AQA applications require distinct modalities (e.g., audio in sports \cite{xia2023skating}, text in action feedback \cite{zhang2024narrative}, and pose in rehabilitation \cite{bruce2024egcn++,10049714}), challenging the generalizability of fixed-architecture multimodal models. Therefore, a flexible framework for incomplete multimodal scenarios is crucial.
    
    Prior solutions address incomplete modalities problems by modality completion or cross-modal joint representation learning. The former reconstructs missing data using available ones with complex generators like diffusion models \cite{wang2023incomplete,meng2024multi}, variational autoencoders \cite{shi2019variational,wu2018multimodal}, and generative adversarial networks \cite{cai2018deep,yoon2018gain}, which incur high computational costs unsuitable for real-time scenes. In contrast, the latter extracts joint cross-modal features \cite{xu2024leveraging,park2023cross} at a lower cost. This joint learning is also widely used for multimodal learning \cite{tip/ZengZ24,xu2024vision} and has received more attention. However, state-of-the-art methods \cite{tip/ZengZ24,xu2024leveraging,park2023cross} often adopt costly two-stage pipelines that sequentially learn unimodal and cross-modal representations (Fig. \ref{fig:1} (a)). This inevitably increases training and optimization costs. Thus, efficiently integrating specific and shared modality knowledge to compensate for missing modalities is of great significance.
    
    While seeking efficient solutions, we identify an indispensable element: the capability to adaptively model correlations between available and missing modalities. Adaptive cross-modal fusion reduces reliance on high-fidelity reconstruction, minimizing dependence on heavyweight generators. Also, specific unimodal knowledge drives precise integration of reconstructed and available modalities. Thus, we naturally turn to the prevalent \textbf{M}ixture \textbf{o}f \textbf{E}xperts (\textbf{MoE})~\cite{li2025ipcmoe}, which flexibly employs experts specialized in processing diverse modal inputs. Based on the benefits of MoE, unimodal experts facilitate accurate modality completion, while selective collaboration specializes in diverse incomplete combinations. By exploiting complementarity between \textbf{M}issing \textbf{M}odality \textbf{C}ompletion (\textbf{MMC}) and MoE, cross-modal knowledge dynamically refines features generated by MMC, enhancing incomplete multimodal learning.
    
    To achieve this, we propose a novel \textbf{M}issing \textbf{C}ompletion Framework with \textbf{M}ixture \textbf{o}f \textbf{E}xperts (\textbf{MCMoE}), unifying unimodal and joint learning in single-stage training. Specifically, we adaptively generate missing modalities from all available ones to preserve modality-specific knowledge modeling despite incompleteness. Then, we learn unimodal experts for each modality and design a soft router to dynamically fuse semantics, compensating for generated features and reducing reliance on heavyweight generators. By aligning the complete modality-specific and generated cross-modal representations, the model is motivated to focus on both specific and shared knowledge modeling. As shown in Fig. \ref{fig:1} (b), our MCMoE leverages the complementarity between MMC and MoE, achieving a balanced learning of unimodal and joint representations within single-stage training.
    
    To bridge cross-modal semantic gaps, we further design a shared temporal enhancement module that alleviates the difficulty of completing missing data from available ones. Then, we propose a novel \textbf{A}daptive \textbf{G}ated \textbf{M}odality \textbf{G}enerator (\textbf{AGMG}) that cost-effectively adapts to various incomplete combinations. AGMG dynamically processes existing modalities to iteratively complete missing ones and employs gating layers for selective fusion. Finally, a cross-modal fusion module integrates the modality features processed by MoE for quality assessment. Extensive experiments on three public AQA benchmarks (Rhythmic Gymnastics~\cite{zeng2020hybrid}, Fis-V~\cite{16}, and FS1000~\cite{xia2023skating}) show that our MCMoE outperforms state-of-the-art methods in both complete and incomplete multimodal scenarios. Abundant ablations validate the contribution of each component. \emph{To our knowledge, this is the first work to explore incomplete multimodal action quality assessment.} The main contributions are:
    \begin{itemize}
        \item We propose a novel missing completion framework with mixture of experts for incomplete multimodal action quality assessment, reducing the reliance on heavyweight generative architectures with unified unimodal and joint representation learning in single-stage training.
        \item We propose the adaptive gated modality generator to selectively complete missing modalities from available ones, and design shared temporal enhancement and cross-modal fusion modules to fill the semantic gaps between modalities and fuse cross-modal semantics.
        \item We conduct extensive experiments and ablation studies to reveal the complementarity between missing modality generation and mixture of experts and the state-of-the-art performance of our method in both complete and incomplete multimodal scenarios on three public benchmarks.
    \end{itemize}
    \begin{figure*}[t]
        \centering
        \includegraphics[width=0.95\linewidth]{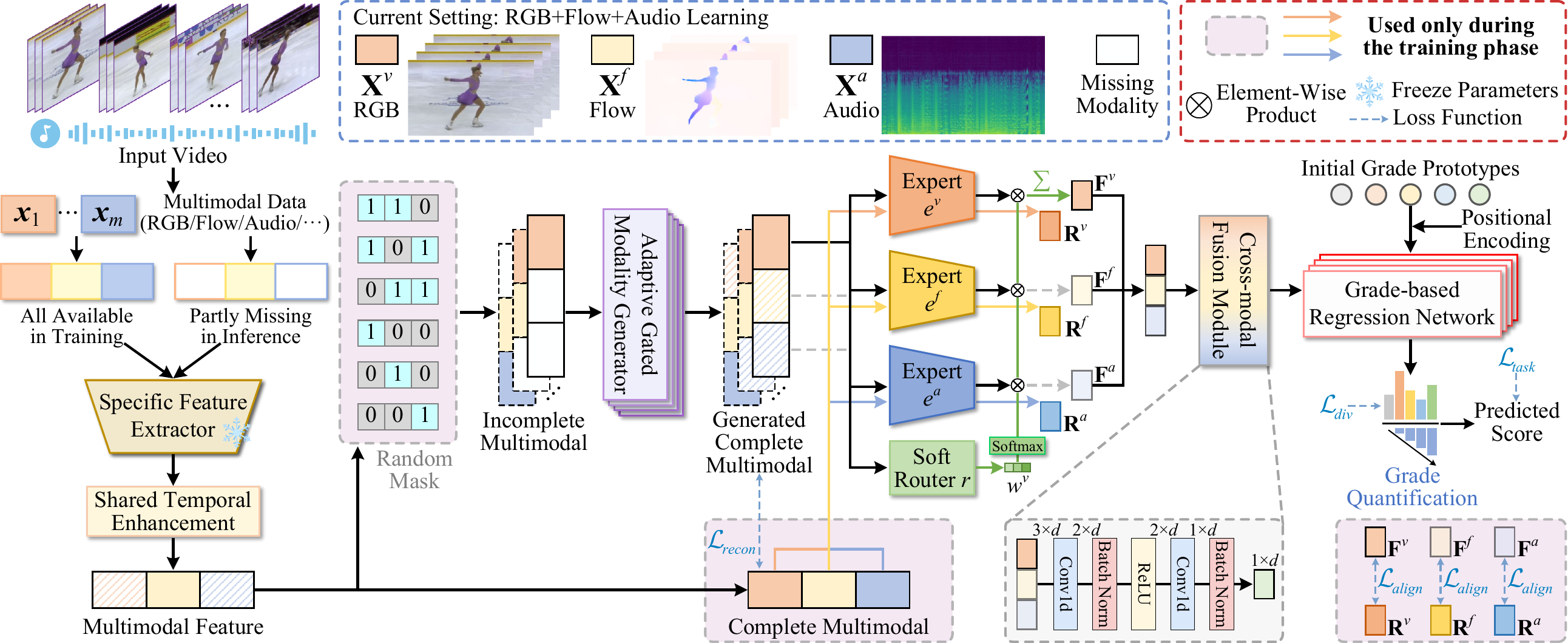}
        \caption{Overview of our missing completion framework with mixture of experts (MCMoE). Following the SOTA multimodal AQA method PAMFN, we use RGB, Flow, and Audio inputs. All modalities are visible during training, and the missing inputs during inference are zero-vector initialized. Frozen modality-specific extractors extract features, enhanced by a shared temporal enhancement module to bridge cross-modal gaps. Random masking simulates modality incompleteness during training and an adaptive gated modality generator completes missing representations. Then, unimodal experts and a soft router enable dynamic fusion, followed by cross-modal integration and grade-based regression for score prediction. (Best viewed in color.)}
        \label{fig:2}
    \end{figure*}
    \section{Related Work}
    \subsection{Multimodal Action Quality Assessment}
    Multimodal learning~\cite{11164481,cai2024efficient,cai2025keep,wu2025enhanced,DBLP:conf/ijcai/HuangSXK25} has achieved notable success recently. Similarly, multimodal AQA aims to evaluate action execution by integrating complementary cues (e.g., RGB, flow, audio, text), offering finer semantic understanding than unimodal methods \cite{xu2022finediving,zhou2023hierarchical}. Recent works \cite{xu2024vision,majeedi2024rica,xia2023skating,tip/ZengZ24} have explored multi-granular semantic alignment, temporal visual-audio fusion, and adaptive modality integration. However, they assume full modality availability during both training and inference, which rarely holds in real-world scenarios due to sensor failure \cite{liu2021face}, environment \cite{12}, privacy concerns \cite{jaiswal2020privacy} or upper-layer algorithm failure \cite{DBLP:journals/iotj/WuZLCJ25,DBLP:conf/mm/Wu0YHFJYL24,DBLP:journals/tdsc/WuLZZBZR24,chen2025causal}. Moreover, modality importance varies across domains, e.g., audio in sports \cite{xia2023skating}, text in feedback \cite{zhang2024narrative}, and gesture in rehabilitation \cite{bruce2024egcn++}. This poses higher challenges to the modal adaptability of models. \emph{Thus, we address this gap by introducing the first framework for incomplete multimodal AQA, adaptable to diverse missing-modality conditions.}
    
    \subsection{Incomplete Multimodal Learning}
    Incomplete multimodal learning has become important for real-world tasks such as emotion recognition, action recognition, and medical analysis \cite{xu2024leveraging,park2023cross,shi2024passion}. Existing solutions mainly rely on (1) modality completion via generative models (e.g., VAE \cite{shi2019variational}, GAN \cite{yoon2018gain}, Diffusion \cite{meng2024multi}) or (2) joint representation learning through cross-modal consistency \cite{lian2023gcnet,park2023cross,xu2024leveraging}. Yet the former depends on heavy generative models, and the latter often requires two-stage training, both increasing cost and complexity. \emph{In contrast, our method exploits the complementarity between modality completion and mixture of experts to unify unimodal and joint learning in a single stage.}
    
    \section{Method}
    In this section, we detail our MCMoE, which combines modality completion with MoE to jointly learn unimodal and joint representations in single-stage training (Fig. \ref{fig:2}).
    \subsection{Overview}
    Our framework accepts multimodal inputs. Its flexible architecture readily extends to arbitrary modality counts. Following mainstream incomplete multimodal research \cite{xu2024leveraging,lian2023gcnet,woo2023towards} and the SOTA multimodal AQA method \cite{tip/ZengZ24}, we focus on three modalities: RGB visual ($v$), optical flow ($f$), and audio ($a$). Following convention, videos are divided into $T$ segments, which are fed into pre-trained modality-specific extractors to obtain features. These features are processed by a shared temporal enhancement module to bridge cross-modal semantic gaps, yielding a multimodal feature set ${\mathbf{X}^M}\!=\!\{ {{\mathbf{X}^m}|m \in \{ {v,f,a} \}} \}$. For each modality, the unimodal set is ${\mathbf{X}^m}\!=\!\{ {\boldsymbol{x}_t^m}\}_{t = 1}^T$. Under incomplete modalities, ${\mathbf{X}^M}$ partitions into available (${{\mathbf{X}}^{\tilde M}}\!=\!\{ {{\mathbf{X}^{\tilde m}}|\tilde m \in \tilde M}\}$) and missing (${{\mathbf{X}}^{\bar M}}\!=\!\{ {{\mathbf{X}^{\bar m}}|\bar m \in \bar M} \}$) subsets, where $M\!=\!\tilde M \cup \bar M\!=\!\left\{ {v,f,a} \right\}$ and $\tilde M \cap \bar M\!=\!\emptyset$.

    We propose an adaptive gated modality generator $\mathcal G$ to dynamically complete missing features ${{\mathbf{X}}^{\bar M}}$ using available features ${{\mathbf{X}}^{\tilde M}}$, optimized via reconstruction loss ${{\mathcal L}_{recon}}$. Generated features ${{\hat {\mathbf{X}}}^{\bar M}}$ combine with ${{\mathbf{X}}^{\tilde M}}$ to form a new complete feature set ${{\hat {\mathbf{X}}}^{M}}\!=\!\{ {{{\hat {\mathbf{X}}}^m}|m \in M} \}$. We design a mixture-of-experts ${\mathcal E^M}\!=\!\left\{ {e^m}|m \in M \right\}$ to dynamically extract both unimodal and joint representations: expert $e^m$ mines modality-specific knowledge from ${{\hat {\mathbf{X}}}^{m}}$, while others assist in capturing cross-modal joint patterns. Then, a shared soft router $r$ selectively fuses these features. Formally,
    \begin{equation}
        \small
        {{\hat {\mathbf{X}}}^M} = {\mathbf{X}^{\tilde M}} \cup {{\hat {\mathbf{X}}}^{\bar M}},{{\hat {\mathbf{X}}}^{\bar M}} = {\mathcal G}\left( {\left( {{\mathbf{X}^{\bar M}},{\mathbf{X}}^{\tilde M}} \right)|\psi } \right),
    \end{equation}
    \begin{equation}
        \small
        {\mathbf{F}^m} = \sum\limits_{k \in \left\{ {v,f,a} \right\}} {r\left( {{{\hat {\mathbf{X}}}^m}|\sigma } \right){e^k}} \left( {{{\hat {\mathbf{X}}}^m}|{\tau ^k}} \right),
    \end{equation}
    where $\psi$, $\sigma$, and $\tau ^k$ denote the learnable parameters of $\mathcal G$, $r$, and expert ${e ^k}$ of modality $k$. ${\mathbf{F}^m}$ is the feature of modality $m$ that fuses modality-specific and shared information.

    During training, we obtain the unimodal feature ${\mathbf{R}^m} = {e^m}( {{{\hat {\mathbf{X}}}^m}|{\tau ^m}})$ using the modality expert $e^m$. An alignment loss ${{\mathcal L}_{align}}$ between ${\mathbf{R}^m}$ and ${\mathbf{F}^m}$ motivates joint learning of unimodal and joint representations. Then, all ${\mathbf{F}^M}$ are concatenated and processed by the cross-modal fusion module $\mathcal{C}$ to capture complementary patterns, yielding multimodal features $\mathbf{H}$. Finally, a grade-based regression network $\mathcal R$ models performance quality patterns ${\mathbf{P}^N}$ from $\mathbf{H}$, which a fully connected layer regresses to rank weights ${\boldsymbol{s}^N}$. The final score $s$ combines ${\boldsymbol{s}^N}$ with grade quantifications ${\mathbf{G}^N}$, where $N$ is the number of grades. Formally,
    \begin{equation}
        \small
        \mathbf{H} = {\mathcal C}\left( {{\rm{Concat}}\left( {{\mathbf{F}^M}} \right)|\phi } \right),
        \label{eq3}
    \end{equation}
    \begin{equation}
        \small
        s = \sum\limits_{n = 1}^N {{\boldsymbol{s}^n} \otimes } {\mathbf{G}^n},{\mathbf{P}^N} = {\mathcal R}\left( {\mathbf{H}|\varphi } \right),
        \label{eq4}
    \end{equation}
    where $\phi$, $\varphi$ are learnable parameters of $\mathcal C$ and $\mathcal R$, and $\otimes$ denotes element-wise product in a batch. A diversity loss ${{\mathcal L}_{div}}$ ensures grade patterns focus on distinct performance aspects, while task-specific loss ${{\mathcal L}_{task}}$ fits quality assessment. With score label $\hat{s}$, the final objective $\mathcal{J}$ is:
    \begin{equation}
        \small
        \begin{split}
            \min~\mathcal{J} &= {\lambda _1}{\mathcal{L}_{recon}}({{\hat {\mathbf{X}}}^M},{\mathbf{X}^M}) + {\lambda _2}{\mathcal{L}_{align}}({\mathbf{R}^M},{\mathbf{F}^M}) \\
            & + {\lambda _3}{\mathcal{L}_{div}}({\mathbf{P}^N}) + {\lambda _4}{\mathcal{L}_{task}}(s,\hat s),
        \end{split}
        \label{eq5}
    \end{equation}
    where $\lambda _1$, $\lambda _2$, $\lambda _3$, and $\lambda _4$ are the balancing weights.
    \subsection{Feature Extraction}
    For fairness, we follow preprocessing pipelines from existing multimodal AQA works~\cite{tip/ZengZ24,xia2023skating}. On Rhythmic Gymnastics and Fis-V, videos are split into $T$ non-overlapping 32-frame segments $\mathbf{I}_T$. Within each segment, frozen pre-trained extractors---VST~\cite{liu2022video} for RGB, I3D~\cite{i3d} for flow, and AST~\cite{gong2021ast} for audio---extract temporally aligned features $\mathbf{I}_T^v$, $\mathbf{I}_T^f$, and $\mathbf{I}_T^a$ of dimension 1024, 1024, and 768, respectively. For FS1000, segments are 5 seconds long with 3 seconds overlap, and TimeSformer~\cite{gberta_2021_ICML}, I3D, and AST process non-overlapping 8-frame clips in each segment with dimensions 768, 1024, and 768. Formally,
    \begin{equation}
        \small
        \mathbf{I}_T^v = \mathcal{V}\left( {{\mathbf{I}_T}} \right),\mathbf{I}_T^f = \mathcal{F}\left( {{\mathbf{I}_T}} \right),\mathbf{I}_T^a = \mathcal{A}\left( {{\mathbf{I}_T}} \right),
    \end{equation}
    where $\mathcal{V}$, $\mathcal{F}$, and $\mathcal{A}$ are the frozen specific feature extractors.

    For cross-modal interactions, inputs are projected to a shared latent space of dimension $d$ via modality-specific modules $\mathcal{P}^M$ with parameters ${{\omega ^M}}$. We then design a \textbf{S}hared \textbf{T}emporal \textbf{E}nhancement \textbf{M}odule (\textbf{STEM}) $\mathcal T$ to bridge semantic gaps. Implemented as stackable Transformer encoders \cite{vaswani2017attention}, $\mathcal{T}$'s shared parameters $\vartheta$ capture cross-modal commonalities while enhancing modalities separately. This avoids direct multimodal feature interaction, facilitating subsequent unimodal learning. Formally,
    \begin{equation}
        \small
        {\mathbf{X}^M} = \mathcal{T}\left( {{\mathcal{P}^M}\left( {\mathbf{I}_T^M|{\omega ^M}} \right)|\vartheta } \right).
    \end{equation}

    \subsection{Adaptive Gated Modality Generator}
\begin{figure}[t]
        \centering
        \includegraphics[width=1.0\linewidth]{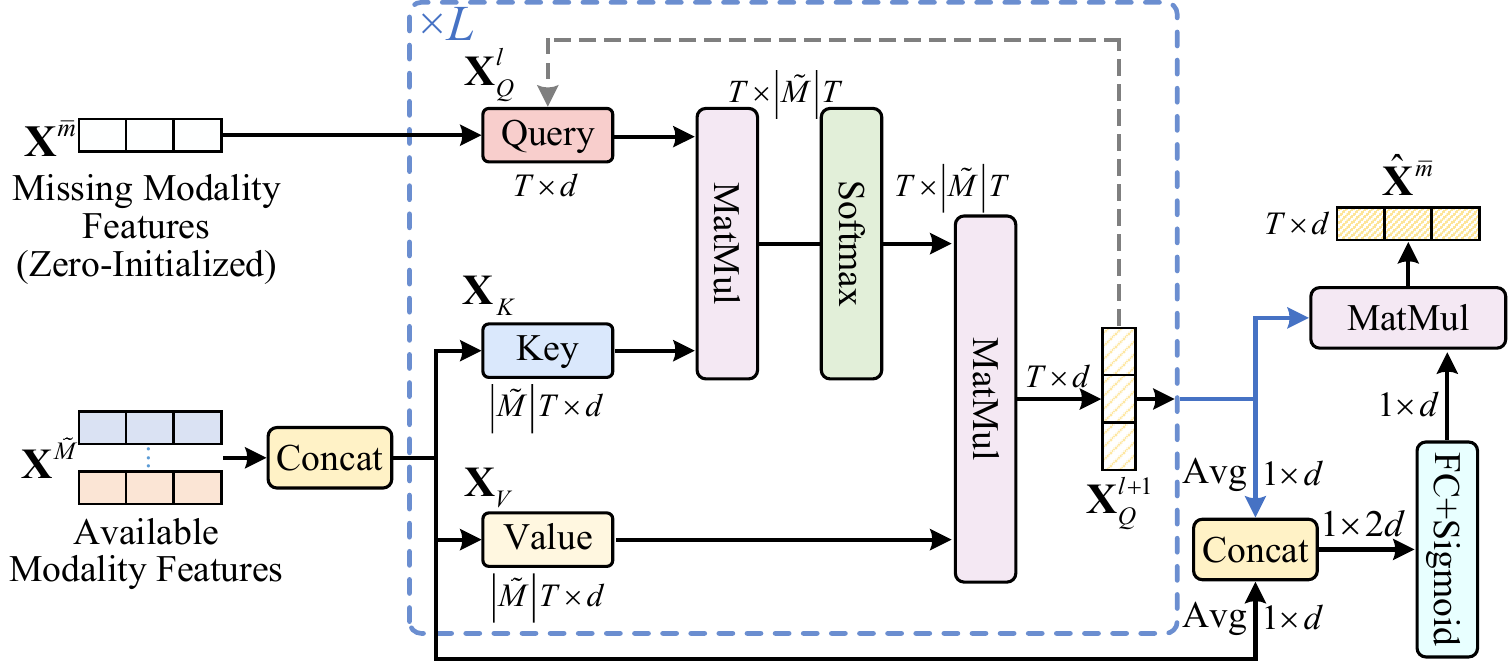}
        \caption{The illustration of our proposed Adaptive Gated Modality Generator (AGMG).}
        \label{fig:3}
    \end{figure}

    To complete the missing modalities, we propose a novel \textbf{A}daptive \textbf{G}ated \textbf{M}odality \textbf{G}enerator (\textbf{AGMG}), which adaptively generates absent features iteratively based on available ones (Fig.~\ref{fig:3}). Inspired by \cite{vaswani2017attention}, AGMG first applies multi-head cross-attention: the concatenated available modality features ${\mathbf{X}}^{\tilde M}$ serve as \emph{keys/values}, while zero-initialized missing features ${{\mathbf{X}}^{\bar M}}$ act as \emph{queries}. Outputs become subsequent layer queries for iterative refinement over $L$ layers ($l$ denotes the current $l$-th layer). Thus, for the missing modality ${\bar m}$, $\mathbf{X}_Q^l = \mathbf{W}_Q^{\bar m}{\mathbf{X}^{\bar m}}$,
    \begin{equation}
        \small
        {\mathbf{X}_K} = \mathbf{W}_K^{\bar m}{\text{Concat}}\left( {{{\mathbf{X}}^{\tilde M}}} \right),{\mathbf{X}_V} = \mathbf{W}_V^{\bar m}{\text{Concat}}\left( {{{\mathbf{X}}^{\tilde M}}} \right),
    \end{equation}
    \begin{equation}
        \small
        \mathbf{X}_Q^{l + 1} = {\text{Softmax}}\left( {\mathbf{X}_Q^l{{\left( {{\mathbf{X}_K}} \right)}^{\rm T}}/\sqrt d } \right){\mathbf{X}_V},
    \end{equation}
    where $\sqrt d$ is a normalization factor, $\mathbf{W}_Q^{\bar m}$, $\mathbf{W}_K^{\bar m}$, and $\mathbf{W}_V^{\bar m}$ are modality-specific learnable weights. The final layer yields $\mathbf{X}_{new}^{\bar m}$. AGMG then employs gating layers $g$ to dynamically weight fusions based on current input completeness, mitigating potential error propagation from imperfect generation:
    \begin{equation}
        \small
        {{\hat {\mathbf{X}}}^{\bar m}} = \mathbf{X}_{new}^{\bar m}\! \cdot\! g\left( {{\text{Avg}}\left( {{\text{Concat}}\left( {{{\mathbf{X}}^{\tilde M}}} \right)} \right),{\text{Avg}}\left( {\mathbf{X}_{new}^{\bar m}} \right)} \right).
    \end{equation}

    In inference, missing modalities are zero-initialized without any information leakage, as in \cite{xu2024leveraging,woo2023towards}. In training, we simulate incompleteness via random masking of complete modalities.

    \begin{table*}[!t]
        \centering
        \small
        \tabcolsep=1.35pt
        \renewcommand\arraystretch{0.9}
        \begin{tabular}{@{}c|c|c|cccccccc@{}}
            \toprule
            \multirow{2}{*}{Datasets}   & \multirow{2}{*}{Methods} & \multirow{2}{*}{Year}  & \multicolumn{8}{c}{Testing Condition (Spearman Correlation (↑)/Mean Square Error (↓))}                                                                                                  \\ \cmidrule(l{0pt}r{0pt}){4-11}
            &               &           & $\{v,f\}$          & $\{v,a\}$           & $\{f,a\}$           & $\{v\}$             & $\{f\}$            & $\{a\}$             & Average           & $\{v,f,a\}$        \\ \midrule
            \multirow{8}{*}{\begin{tabular}[c]{@{}c@{}}FS1000\\      (7-class)\end{tabular}} & $\heartsuit$MLP-Mixer*    & 2023            & 0.722/25.56  & 0.542/60.57     & 0.474/87.20     & 0.623/101.35    & 0.472/87.59    & 0.177/68.43     & 0.520/71.78     & 0.819/14.56    \\
            & $\heartsuit$PAMFN*    & 2024               & 0.727/59.96    & 0.644/62.80     & 0.561/56.36     & 0.713/92.10     & 0.486/117.63   & 0.145/62.13     & 0.571/75.16     & \underline{0.855}/\underline{13.02}    \\
            & $\clubsuit$ActionMAE*   & 2023            & 0.775/24.66    & 0.766/64.13     & 0.556/26.51     & 0.761/50.64     & 0.462/{\textbf{21.47}}    & 0.458/41.66     & 0.651/38.18     & 0.809/17.96    \\
            & $\spadesuit$GCNet*     & 2023              & 0.730/25.56    & 0.740/\underline{23.86}     & 0.507/24.97     & 0.696/\underline{26.67}     & 0.447/31.27    & 0.442/39.40     & 0.610/28.62     & 0.764/21.82    \\
            & $\spadesuit$IMDer*       & 2023            & 0.760/22.34    & 0.745/28.46     & \underline{0.573}/24.86    & 0.724/35.99     & 0.424/22.92    & 0.488/32.56     & 0.636/27.86     & 0.788/25.95    \\
            & $\spadesuit$MoMKE*       & 2024           & \underline{0.798}/18.86    & \underline{0.805}/23.88     & 0.541/24.96     & \underline{0.785}/37.96     & 0.398/23.31    & \underline{0.499}/27.53     & \underline{0.668}/26.08     & 0.819/16.85    \\
            & $\spadesuit$SDR-GNN*         & 2025           & 0.789/\underline{17.50}    & 0.785/25.08     & 0.564/\underline{22.29}     & 0.749/28.47     & \underline{0.504}/29.96    & 0.477/\underline{25.46}     & 0.665/\underline{24.79}     & 0.817/15.91    \\
            & \textbf{MCMoE (Ours)}   & -          & {\textbf{0.845}}/{\textbf{12.66}}    & {\textbf{0.882}}/{\textbf{11.85}}     & {\textbf{0.738}}/{\textbf{14.88}}     & {\textbf{0.845}}/{\textbf{13.64}}     & {\textbf{0.650}}/\underline{22.47}    & {\textbf{0.615}}/{\textbf{16.72}}     & {\textbf{0.782}}/{\textbf{15.37}}     & {\textbf{0.881}}/{\textbf{11.53}}    \\ \midrule
            \multirow{8}{*}{\begin{tabular}[c]{@{}c@{}}Fis-V\\      (2-class)\end{tabular}}    & $\heartsuit$MLP-Mixer*     & 2023           & 0.732/30.34    & 0.651/46.70     & 0.572/26.96     & 0.618/48.46     & 0.546/27.18    & 0.325/67.25     & 0.586/41.15     & 0.772/\underline{13.97}    \\
            & $\heartsuit$PAMFN         & 2024           & \underline{0.801}/33.49    & 0.661/54.34     & 0.622/110.50    & 0.644/84.93     & 0.616/110.42   & 0.141/86.16     & 0.610/79.97     & \underline{0.822}/15.33    \\
            & $\clubsuit$ActionMAE*       & 2023         & 0.704/33.54    & 0.678/27.61     & 0.575/25.55     & 0.616/40.07     & 0.484/24.87    & 0.486/29.29     & 0.597/30.16     & 0.698/17.34    \\
            & $\spadesuit$GCNet*        & 2023            & 0.738/19.86    & 0.656/21.32     & 0.594/21.72     & 0.667/20.87     & 0.602/\underline{19.61}    & 0.455/34.77     & 0.626/23.03     & 0.698/16.93    \\
            & $\spadesuit$IMDer*       & 2023             & 0.748/15.19    & 0.658/22.66     & 0.568/23.99     & 0.675/25.45     & 0.618/26.38    & 0.405/31.96     & 0.622/24.27     & 0.703/17.02    \\
            & $\spadesuit$MoMKE*      & 2024              & 0.754/\underline{14.84}    & \underline{0.689}/\underline{20.60}     & \underline{0.646}/\underline{19.62}     & \underline{0.684}/23.16     & \underline{0.654}/22.46    & \underline{0.497}/\underline{29.09}     & \underline{0.660}/\underline{21.63}     & 0.747/17.30    \\
            & $\spadesuit$SDR-GNN*        & 2025           & 0.752/14.99    & 0.680/20.62     & 0.619/20.89     & 0.689/\underline{20.26}     & 0.648/21.50    & 0.479/32.13     & 0.651/21.73     & 0.733/16.45    \\
            & \textbf{MCMoE (Ours)}     & -        & {\textbf{0.813}}/{\textbf{11.02}}    & {\textbf{0.787}}/{\textbf{14.64}}     & {\textbf{0.727}}/{\textbf{17.41}}     & {\textbf{0.765}}/{\textbf{15.14}}     & {\textbf{0.698}}/{\textbf{15.39}}    & {\textbf{0.557}}/{\textbf{28.54}}     & {\textbf{0.734}}/{\textbf{17.02}}     & {\textbf{0.829}}/{\textbf{12.15}}    \\ \midrule
            \multirow{8}{*}{\begin{tabular}[c]{@{}c@{}}RG\\      (4-class)\end{tabular}}      & $\heartsuit$MLP-Mixer*      & 2023          & 0.733/7.23     & 0.614/14.09     & 0.485/10.18     & 0.655/9.67      & 0.566/11.45    & 0.244/16.01     & 0.567/11.44     & 0.754/7.48     \\
            & $\heartsuit$PAMFN        & 2024            & \underline{0.764}/6.78     & 0.616/38.87     & 0.448/122.11    & 0.658/39.86     & 0.483/123.44   & 0.131/151.69    & 0.543/80.46     & \underline{0.819}/6.64     \\
            & $\clubsuit$ActionMAE*       & 2023         & 0.724/7.30     & 0.621/8.76      & 0.545/10.39     & 0.689/12.41     & 0.521/11.29    & 0.251/16.84     & 0.575/11.16     & 0.709/7.01     \\
            & $\spadesuit$GCNet*          & 2023          & 0.738/6.69     & 0.638/7.95      & 0.556/11.75     & 0.701/8.12      & 0.568/36.01    & 0.225/15.20     & 0.591/14.29     & 0.716/6.45     \\
            & $\spadesuit$IMDer*        & 2023            & 0.746/6.03     & 0.646/7.55      & 0.569/\underline{8.80}      & 0.699/\underline{7.59}      & 0.596/\underline{9.35}     & 0.206/14.19     & 0.598/\underline{8.92}      & 0.724/6.37     \\
            & $\spadesuit$MoMKE*        & 2024            & 0.762/\underline{5.69}     & \underline{0.656}/7.97      & \underline{0.629}/9.39      & 0.693/8.42      & \underline{0.621}/10.08    & \underline{0.264}/\underline{13.25}     & \underline{0.623}/9.13      & 0.747/\underline{6.18}     \\
            & $\spadesuit$SDR-GNN*       & 2025            & 0.758/6.08     & 0.655/\underline{7.38}      & 0.612/9.40      & \underline{0.727}/7.77      & 0.591/9.80     & \underline{0.264}/13.53     & 0.621/8.99      & 0.742/6.35     \\
            & \textbf{MCMoE (Ours)}      & -      & {\textbf{0.822}}/{\textbf{5.33}}     & {\textbf{0.781}}/{\textbf{5.83}}      & {\textbf{0.699}}/{\textbf{8.15}}      & {\textbf{0.767}}/{\textbf{6.25}}      & {\textbf{0.662}}/{\textbf{8.59}}     & {\textbf{0.278}}/{\textbf{13.20}}     & {\textbf{0.697}}/{\textbf{7.89}}      & {\textbf{0.842}}/{\textbf{4.85}}     \\ \bottomrule
        \end{tabular}
        \caption{Comparisons of performance on three benchmarks with incomplete modalities. $v$, $f$, and $a$ refer to the RGB, flow, and audio modalities. ``Average'' denotes the average result of all six incomplete multimodal combinations. The \textbf{bold} / \underline{underline} indicate the best / second-best results. * indicates our reimplementation. $\heartsuit$, $\clubsuit$, and $\spadesuit$ mean the evaluated method sources for multimodal AQA, incomplete multimodal action recognition, and incomplete multimodal emotion recognition.}
        \label{tab:1}
    \end{table*}
    \subsection{Complementarity between MMC and MoE}
    This work exploits the complementarity between \textbf{M}issing \textbf{M}odality \textbf{C}ompletion (\textbf{MMC}) and \textbf{M}ixture \textbf{o}f \textbf{E}xperts (\textbf{MoE}) to avoid heavy generative models and unify unimodal/joint learning in one stage. We detail this in two parts:

    \emph{\textbf{Benefits of MMC for MoE.}} Inspired by prior works \cite{xu2024leveraging,zhang2024mixture}, we employ a MoE to dynamically mix expert knowledge for cross-modal representation in incomplete multimodal scenarios. However, missing modalities limit or mislead cross-modal semantic extraction, and existing two-stage solutions \cite{xu2024leveraging,park2023cross} increase training cost and optimization complexity. In contrast, MMC generates higher-confidence features instead of zero matrices based on available modalities. Guided by the reconstruction loss $\mathcal L_{recon}$ with real-modality supervision, these features carry reliable unimodal knowledge and cross-modal cues, benefiting both unimodal and joint learning for modality experts and enabling experts with fewer parameters.

    \emph{\textbf{Benefits of MoE for MMC.}} Existing MMC methods often rely on heavyweight generative models \cite{cai2018deep,wang2023incomplete,meng2024multi} to complete missing modalities. In contrast, our MoE dynamically selects and fuses modality knowledge to handle generated features. It adaptively balances real available and generated features to prevent error propagation from imperfect generation. When processing missing modality $\bar m$, MoE further refines missing data via inter-modal correlations from available ones, boosting robustness and accuracy of joint representations. These advantages reduce reliance on high-fidelity generators in MMC and lower parameter costs.

    Building on the above complementarity, our MoE adopts a lightweight two-layer multilayer perceptron (MLP) for each expert. We aggregate all experts' outputs to derive both unimodal and joint representations for a given modality. For example, the unimodal ($\mathbf{F}_v^v$) and joint representations ($\mathbf{F}_f^v$ and $\mathbf{F}_a^v$) of visual modality $v$ can be obtained as follows:
    \begin{equation}
        \small
        \mathbf{F}_m^v = {e^m}\left( {{{\hat {\mathbf{X}}}^v}} \right) = {\text{ML}}{{\text{P}}_{{\tau ^m}}}\left( {{{\hat {\mathbf{X}}}^v}} \right),m \in \left\{ {v,f,a} \right\}.
    \end{equation}

    To handle diverse missing-modality scenarios, we employ a two-layer MLP soft router $r$ to dynamically estimates the importance of unimodal and joint representations based on the input ${{{\hat {\mathbf{X}}}^v}}$. The final visual feature $\mathbf{F}^v$ is obtained by weighting based on the importance weights $\boldsymbol{w}_m^v$:
    \begin{equation}
        \small
        \left\{ {\boldsymbol{w}_v^v,\boldsymbol{w}_f^v,\boldsymbol{w}_a^v} \right\} = r\left( {{{\hat {\mathbf{X}}}^v}} \right) = {\text{Softmax}}\left( {{\text{ML}}{{\text{P}}_\sigma }\left( {{{\hat {\mathbf{X}}}^v}} \right)} \right),
    \end{equation}
    \begin{equation}
        \small
        {\mathbf{F}^v} = \sum\limits_{m \in \left\{ {v,f,a} \right\}} {\boldsymbol{w}_m^v \cdot } \mathbf{F}_m^v.
    \end{equation}

    The flow ($\mathbf{F}^f$) and audio ($\mathbf{F}^a$) features are obtained similarly. We also apply the alignment loss $\mathcal L_{align}$ to align $\mathbf{F}^m$ with true unimodal features ${\mathbf{R}^m}\!=\!{e^m}({{\hat {\mathbf{X}}}^m})$. This motivates both unimodal and joint learning within single-stage training, and also indirectly promotes AGMG to generate more faithful information. Finally, we design a \textbf{C}ross-modal \textbf{F}usion \textbf{M}odule (\textbf{CFM}) to capture inter-modal correlations and map features into a task-specific latent space. Our CFM can rely on a simple convolutional block (Fig.~\ref{fig:2}) to leverage the MoE's cross-modal semantics for efficient fusion.

    \subsection{Score Generation and Optimization}
    As shown in Eq. \ref{eq3}, we extract multimodal features $\mathbf{H}$ via CFM. For scoring, we employ state-of-the-art grade-based regression \cite{xu2022likert,Xu_2025_CVPR,10884538}. We initialize $N$ learnable grade prototypes with sine–cosine positional encodings, then use a three-layer Transformer decoder to implement regressor $\mathcal R$ to aggregate $\mathbf{H}$ into grade patterns ${\mathbf{P}^N}$. The final score $s$ is computed by Eq.~\ref{eq4}, where the grade quantification $\mathbf{G}^n = \frac{{n - 1}}{{N - 1}}$.

    As shown in Eq. \ref{eq5}, our method uses four losses: reconstruction loss ${{\mathcal L}_{recon}}$, alignment loss ${{\mathcal L}_{align}}$, diversity loss ${{\mathcal L}_{div}}$, and task-specific loss ${{\mathcal L}_{task}}$. Specifically, ${{\mathcal L}_{recon}}$ uses Mean Square Error (MSE) to train AGMG for high-fidelity features. ${{\mathcal L}_{align}}$ minimizes Kullback-Leibler (KL) divergence between $\mathbf{F}^M$ and $\mathbf{R}^M$ for unified unimodal and joint learning. ${{\mathcal L}_{div}}$ applies triplet loss to separate grade patterns. For quality assessment, ${{\mathcal L}_{task}}$ uses MSE to fit predictions to expert scores. Hence, Eq. \ref{eq5} can be rewritten as:
    \begin{equation}
        \small
        {\text{MSE}}\left( {y,\hat y} \right) = {\left\| {y - \hat y} \right\|^2},
        \label{eq16}
    \end{equation}
    \begin{equation}
        \resizebox{.88\linewidth}{!}{$
        {\mathcal{L}_{recon}}\left( {{{\hat {\mathbf{X}}}^M},{\mathbf{X}^M}} \right) = {\text{MSE}}\left( {{{\hat {\mathbf{X}}}^M},{\mathbf{X}^M}} \right),{\mathcal{L}_{task}}(s,\hat s) = {\text{MSE}}\left( {s,\hat s} \right),$}
    \end{equation}
    \begin{equation}
        \resizebox{.88\linewidth}{!}{$
        {\mathcal{L}_{align}}({\mathbf{R}^M},{\mathbf{F}^M}) = {\text{KL}}\left( {{\mathbf{R}^M}\parallel {\mathbf{F}^M}} \right) = \sum\nolimits_t {\mathbf{R}_t^M\log \left( {\frac{{\mathbf{R}_t^M}}{{\mathbf{F}_t^M}}} \right)},$}
    \end{equation}
    \begin{equation}
        \resizebox{.88\linewidth}{!}{$
        {\mathcal{L}_{div}}({\mathbf{P}^N}) = {\sum\limits_n {\left[ {\max \left( {{\text{sim}}\left( {{\mathbf{P}^n},{\mathbf{P}^i}} \right)} \right) - \min \left( {{\text{sim}}\left( {{\mathbf{P}^n},{\mathbf{P}^i}} \right)} \right) + \delta } \right]} _ + },$}
    \end{equation}
    where $i\ne n$, $\delta$ is a margin parameter, which is set to 1. The $\text{sim}\left( \cdot, \cdot  \right)$ denotes cosine similarity, ${{\left[ \cdot  \right]}_{+}}$ means $\max \left( 0,\cdot  \right)$.

    \begin{figure*}[t]
        \centering
        \includegraphics[width=1.0\linewidth]{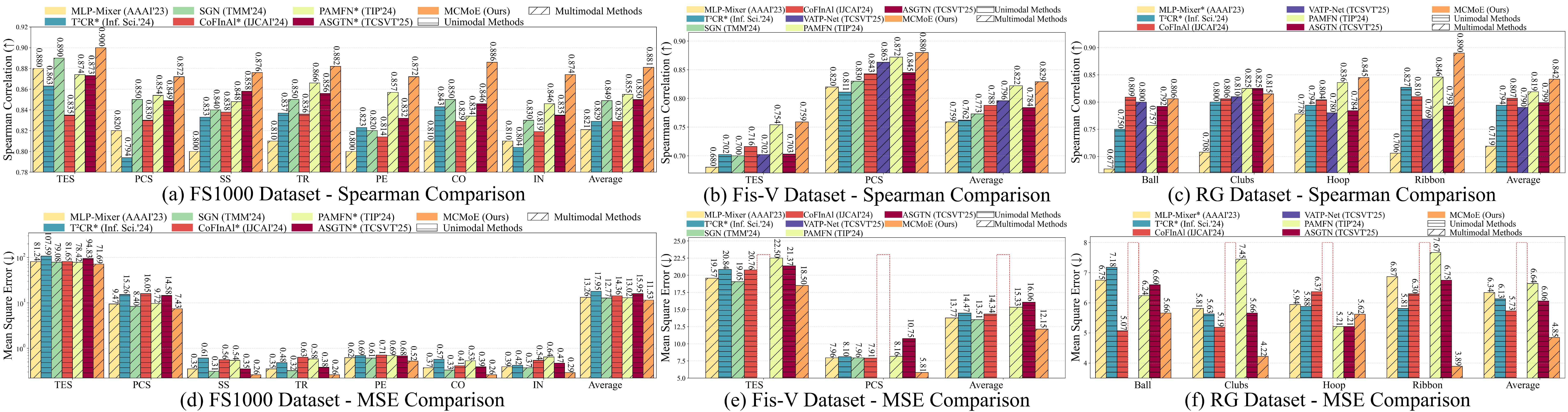}
        \caption{Comparisons of performance with complete modalities. * indicates our reimplementation based on the official code.}
        \label{fig:5}
    \end{figure*}
    \section{Experiments}
    \textbf{Datasets and Metrics.} We evaluate our method on three public AQA benchmarks: FS1000~\cite{xia2023skating}, Fis-V~\cite{16}, and Rhythmic Gymnastics (RG)~\cite{zeng2020hybrid}, which provide RGB, flow, and audio modalities. Following standard protocols \cite{xia2023skating,du2023learning}, we report Spearman’s Rank Correlation ($\rho$) and Mean Square Error (MSE). MSE measures the numerical error (Eq.~\ref{eq16}), while $\rho$ assesses the rank agreement between predictions and ground truth.
    
    \noindent \textbf{Implementation Details.} All experiments use a single RTX 3090 GPU (PyTorch 1.12.0) and 2.40GHz CPU. We evaluate incomplete modalities using the common fixed-missing protocol \cite{xu2024leveraging,wang2023incomplete}, covering the full set $\{v,f,a\}$ and six subsets ($\{v,f\}$, $\{v,a\}$, $\{f,a\}$, $\{v\}$, $\{f\}$, $\{a\}$). For FS1000/Fis-V/RG, we randomly sample 95/124/68 continuous clips. The ${\lambda }_{1}$, ${\lambda }_{2}$, ${\lambda }_{3}$, and ${\lambda }_{4}$ in Eq.~\ref{eq5} are 1, 1, 1/0.5/1, and 10. The grade $N$ is 4. We set a dropout of 0.3 to avoid over-fitting. The batch size is 32 and learning rate is 1e-4/2e-4/2e-4. We optimize with Adam (weight decay 1e-4) and cosine annealing (decay 0.01). For better convergence, we train different models with different epochs as in~\cite{xu2022likert,tip/ZengZ24,CoFInAl}.
    
    \subsection{Comparison with State-of-the-art}
    \textbf{Incomplete Multimodal Scenarios.} To evaluate our method under incomplete modalities, we test all modal combinations on three datasets (Tab. \ref{tab:1}). Comparisons include SOTA methods from multimodal AQA~\cite{xia2023skating,tip/ZengZ24}, incomplete multimodal action recognition~\cite{woo2023towards}, and incomplete multimodal emotion recognition~\cite{lian2023gcnet,wang2023incomplete,xu2024leveraging,fu2025sdr}. We extend the bimodal MLP-Mixer~\cite{xia2023skating} to support trimodal inputs via secondary bimodal interactions. Our MCMoE outperforms all baselines on nearly all metrics and incomplete configurations. Averaged over six combinations, it improves SP. Corr./MSE by 17.1\%/38.0\%, 11.2\%/21.3\%, and 11.9\%/11.5\% on the three datasets. Existing SOTA multimodal AQA methods degrade notably with missing modalities, especially on MSE, likely due to disrupted cross-modal interactions. Incomplete multimodal baselines from other domains lack tailored modeling for action semantics and assessment patterns, resulting in poor AQA performance. Our MCMoE maintains strong performance under both complete and incomplete settings, with average gains of 2.23\%/15.3\% under full modalities. This highlights the benefit of jointly leveraging MMC and MoE to compensate for missing modality interference.
    
    \noindent \textbf{Complete Modality Scenarios.} Fig. \ref{fig:5} compares our method with unimodal\cite{ke2024two,CoFInAl,10884538} and multimodal~\cite{xia2023skating,tip/ZengZ24,gedamu2024visual,du2023learning} SOTA AQA models under full-modality settings. Our method achieves the best or second-best results across all categories and consistently superior averages, with SP. Corr./MSE gains of 3.0\%/9.7\%, 0.9\%/10.1\%, and 2.8\%/15.4\% on the three datasets. This validates the effectiveness of MMC+MoE synergy, which enhances adaptive multimodal fusion and mitigates the impact of missing data. In addition, our single-stage learning avoids knowledge forgetting and complexity issues common in two-stage training.
    
    As shown in Tab. \ref{tab:1} and Fig. \ref{fig:5}, our MCMoE achieves balanced and state-of-the-art performance in both complete and incomplete scenarios. Moreover, Tab.~\ref{tab:4} shows it offers a better performance–efficiency trade-off compared to SOTA incomplete multimodal methods on FS1000.
    
    \begin{table}[t]
        \centering
        \small
        \tabcolsep=1.3pt
        \renewcommand\arraystretch{0.9}
        \begin{tabular}{@{}c|c|c|c|c|c|c@{}}
            \toprule
            Methods               & Year & 1-stage & \#Params & \#FLOPs & Average       & $\{v,f,a\}$     \\ \midrule
            MLP-Mixer           & 2023 & $\checkmark$            & 14.32M        & 49.90G       & 0.52/71.8 & 0.82/14.6 \\
            PAMFN                 & 2024 & $\times$            & 18.06M        & 2.56G        & 0.57/75.2 & 0.86/13.0 \\
            ActionMAE             & 2023 & $\checkmark$            & 14.05M        & 62.12G       & 0.65/38.2 & 0.81/18.0 \\
            GCNet                 & 2023 & $\checkmark$            & 8.78M         & 1191.39G     & 0.61/28.6 & 0.76/21.8 \\
            IMDer                 & 2023 & $\times$            & 7.97M         & 23.53G       & 0.61/27.9 & 0.79/26.0 \\
            MoMKE                 & 2024 & $\times$            & 5.39M         & 2.60G        & 0.67/26.1 & 0.82/16.9 \\
            SDR-GNN           & 2025 &  $\checkmark$           & 22.63M         & 24.35G        & 0.67/24.8 & 0.82/15.9 \\
            \textbf{Ours} & -    & $\checkmark$            & \textbf{4.90M}          & \textbf{1.34 G}        & \textbf{0.78/15.4} & \textbf{0.88/11.5} \\ \bottomrule
        \end{tabular}
        \caption{Compare the computational costs with the SOTA.}
        \label{tab:4}
    \end{table}
    \begin{table}[t]
        \centering
        \small
        \tabcolsep=1.5pt
        \renewcommand\arraystretch{0.9}
        \begin{tabular}{@{}l|ll|ll@{}}
            \toprule
            \multirow{2}{*}{Settings} & \multicolumn{2}{c|}{RG}       & \multicolumn{2}{c}{Fis-V}     \\ \cmidrule(l{0pt}r{0pt}){2-5}
            & Average       & $\{v,f,a\}$    & Average       & $\{v,f,a\}$     \\ \midrule
            Baseline                  & 0.532/25.79 & 0.718/7.05 & 0.577/61.45 & 0.724/16.38 \\
            + STEM                    & 0.573/19.92 & 0.744/6.83 & 0.637/39.72 & 0.748/14.30 \\
            + AGMG                     & 0.647/9.45  & 0.779/6.04 & 0.678/21.36 & 0.773/13.26 \\
            + MoE {\scriptsize(w/o CFM)}           & 0.675/7.91  & 0.817/5.30 & 0.701/18.07 & 0.790/13.10 \\
            \textbf{+ CFM (Ours)}               & \textbf{0.697/7.89}  & \textbf{0.842/4.85} & \textbf{0.734/17.02} & \textbf{0.829}/\textbf{12.15} \\ \midrule
            w/o AGMG                   & 0.635/16.54 & 0.724/6.44 & 0.609/50.45 & 0.742/14.37 \\
            w/o MoE                   & 0.658/7.97  & 0.777/5.48 & 0.684/19.20 & 0.767/13.32 \\
            w/o STEM                  & 0.585/10.96 & 0.735/7.62 & 0.643/19.34 & 0.739/14.16 \\
            w/o ${\mathcal{L}_{recon}}$                & 0.617/9.30  & 0.771/5.50 & 0.699/18.45 & 0.813/12.47 \\
            w/o ${{\mathcal L}_{align}}$                & 0.648/8.56  & 0.797/5.16 & 0.708/19.25 & 0.790/12.59 \\
            w/o ${{\mathcal L}_{div}}$                  & 0.624/8.13  & 0.791/4.93 & 0.693/17.54 & 0.795/12.29 \\ \bottomrule
        \end{tabular}
        \caption{Ablation results on the RG and Fis-V. The top half adds our components in order, and the bottom half individually removes one. Results are shown by $\rho$(↑)/MSE(↓).}
        \label{tab:3}
    \end{table}
    \subsection{Ablation Study}
    To validate the effectiveness of our components, we build a \textbf{Baseline} that extracts and projects multimodal features, directly summed and fed into a grade-based regressor with ${{\mathcal L}{div}}$ and ${{\mathcal L}{task}}$. As shown in Tab. \ref{tab:3}, performance improves as components are incrementally added. Notably, adding AGMG significantly boosts accuracy, showing that dynamic modality completion effectively bridges semantic gaps of missing data. Adding MoE further improves results, especially under incomplete settings, with average gains of 3.9\%/15.9\%. This is likely due to its adaptive fusion of unimodal and cross-modal knowledge. Removing any component from the full model leads to clear performance drops. Notably, STEM is critical for capturing temporal context and cross-modal semantics, essential in long video understanding \cite{xu2022likert,CoFInAl} and multimodal learning \cite{tip/ZengZ24,woo2023towards}.
    
    We also ablate the loss terms. Beyond the core ${\mathcal{L}_{task}}$, removing ${\mathcal{L}_{recon}}$, ${{\mathcal L}_{align}}$, or ${{\mathcal L}_{div}}$ each harms performance. ${\mathcal{L}_{recon}}$ provides key supervision for reliable modality completion, avoiding generating misleading misinformation. ${{\mathcal L}_{align}}$ motivates MoE to jointly learn unimodal and cross-modal features within single-stage training. ${{\mathcal L}_{div}}$ enforces diversity across grade patterns, aiding accurate AQA.
    \begin{figure}[t]
        \centering
        \includegraphics[width=\linewidth]{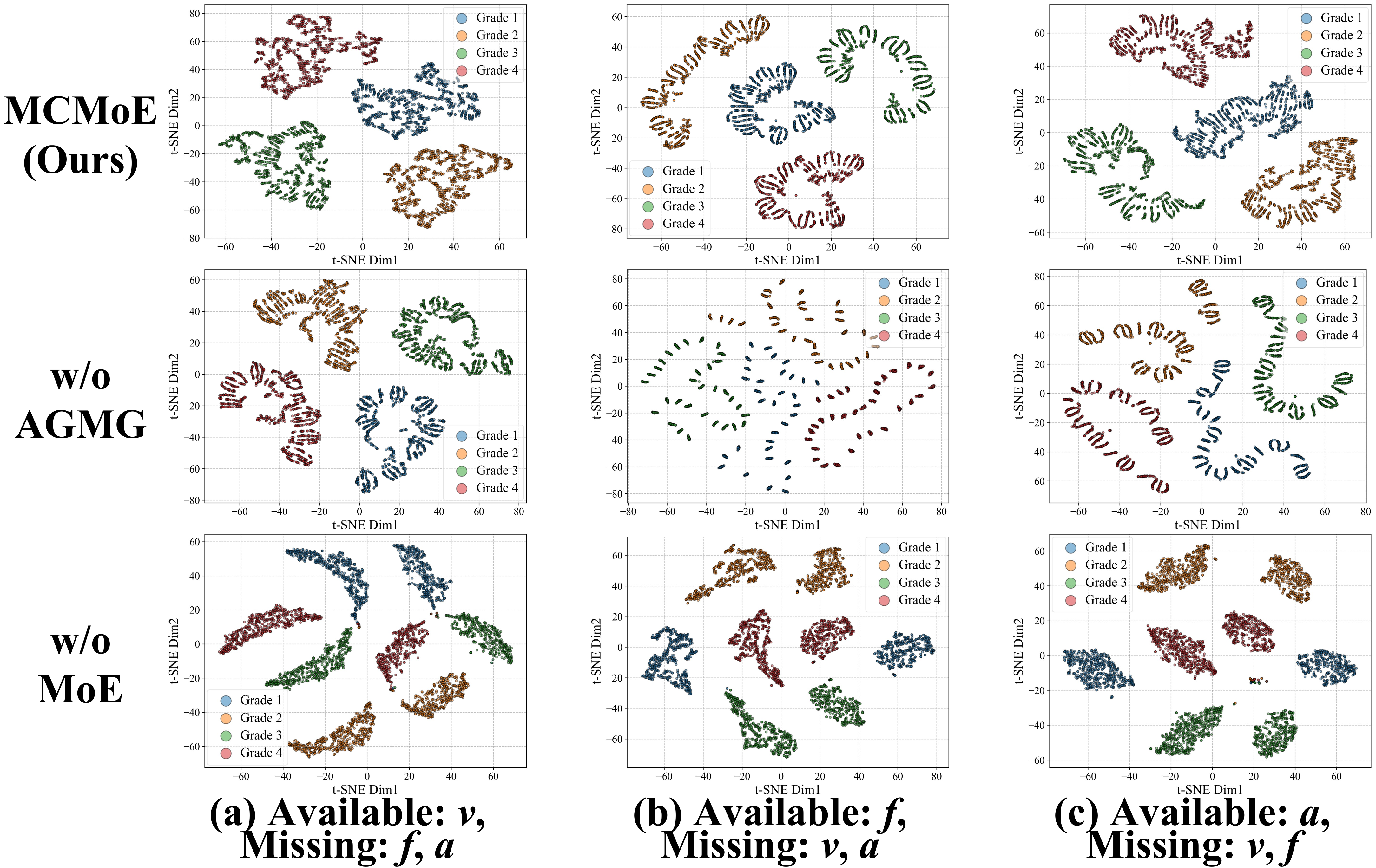}
        \caption{The t-SNE grade distributions in the three extreme unimodal scenes contrasting without AGMG and MoE.}
        \label{fig:4}
    \end{figure}
    \subsection{Visualization Analysis}
    To visualize the complementary effects of our emphasized MMC and MoE, Fig. \ref{fig:4} presents t-SNE distributions of grade patterns on FS1000. Each point is a grade feature. Our MCMoE effectively distinguishes action qualities, yielding compact clusters with clear inter-grade boundaries. In contrast, omitting AGMG or MoE leads to scattered distributions and weaker separability. Without AGMG's modality completion, MoE fails to maintain intra-grade consistency, resulting in dispersed features. Without MoE's dynamic cross-modal fusion, unimodal semantics cannot collaborate effectively, splitting quality grades into two clusters due to modality divergence. This visualization highlights the critical roles of MMC and MoE in quality space modeling.
    
\begin{figure}[t]
        \centering
        \includegraphics[width=\linewidth]{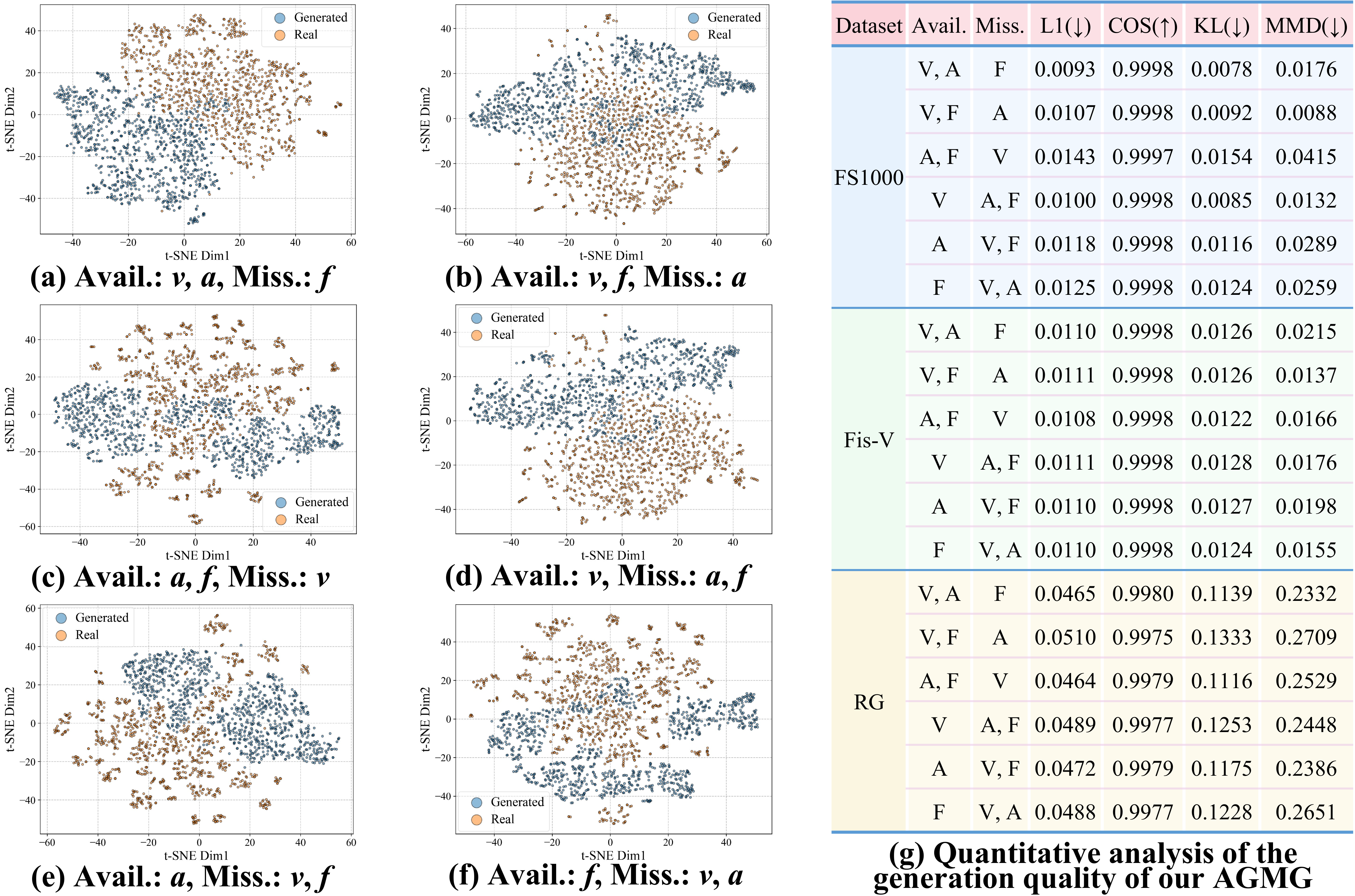}
        \caption{(a-f) t-SNE distributions of our generated and true features; (g) Generation quality analysis on four metrics.}
        \label{fig:6}
    \end{figure}

    Moreover, Fig. \ref{fig:6} qualitatively and quantitatively shows the quality of features generated by AGMG. The generated and real features are well mixed in the t-SNE space in all incomplete combinations on FS1000, indicating high semantic similarity that makes them hard to distinguish. We conduct a comprehensive quantitative analysis using four similarity metrics---L1 Distance, Cosine Similarity, KL Divergence, and Maximum Mean Discrepancy. Even on the smallest RG with limited multimodal training data, AGMG produces high-quality features. These results show that AGMG generates semantically aligned features at a low cost, effectively mitigating the limitations caused by missing data.
    \section{Conclusion}
    In this paper, we introduce MCMoE, a framework for incomplete multimodal action quality assessment that leverages the complementarity between Missing Modality Completion (MMC) and Mixture of Experts (MoE). MCMoE uses an adaptive gated modality generator to reconstruct missing modalities and a MoE architecture with unimodal experts and a soft router to fuse modality-specific and cross-modal information. This design mitigates the impact of missing data, reduces dependence on heavy generative models, and enables unified unimodal and joint representation learning in a single stage. As a result, MCMoE achieves superior performance on three public AQA benchmarks under both complete and incomplete settings, striking a balance between performance and cost.

\section{Acknowledgments}
This work was supported in part by the National Key Research and Development Plan of China under Grant 2021YFB3600503, in part by the National Natural Science Foundation of China under Grant 61972097, U21A20472, 62522102, and 62373043, in part by the Major Scientific Research Project for Technology Promotes Police under Grant 2025YZ040003, 2024YZ040001, in part by the Natural Science Foundation of Fujian Province under Grant 2025J01536.
\appendix
\section{Appendix}
\section{Datasets}
\label{sec:Datasets}
To fully validate the effectiveness of our method, we conduct extensive experiments on three public action quality assessment (AQA) benchmarks. These datasets include two types of sports competitions: figure skating (FS1000~\cite{xia2023skating}) and Fis-V~\cite{16}) and artistic gymnastics (rhythmic gymnastics (RG)~\cite{zeng2020hybrid}).

\noindent\textbf{FS1000.} The FS1000 dataset contains 1,000 training videos and 247 validation videos, representing eight categories of figure skating competitions: men's/ladies'/pairs' short programs, men's/ladies'/ pairs' free skating, and ice dance rhythm/free dances. Each video is recorded at 25 frames per second and contains approximately 5,000 frames. The dataset provides detailed scoring annotations, including Total Element Score (TES), Total Program Component Score (PCS), and five sub-components: Skating Skills (SS), Transitions (TR), Performance (PE), Composition (CO), and Interpretation of Music (IN). As the first figure skating dataset to incorporate audio-visual learning, FS1000 supports rule-compliant multimodal training. Following prior works~\cite{xia2023skating,du2023learning}, we train dedicated models for each scoring category.

\noindent\textbf{Figure Skating Video (Fis-V).} The Fis-V dataset consists of 500 videos of ladies' singles short program performances in figure skating, each approximately 2.9 minutes long and recorded at 25 frames per second. Following the standard split, 400 videos are used for training and 100 for testing. Each video includes official annotations for Total Element Score (TES) and Total Program Component Score (PCS), aligned with competition regulations. Consistent with previous methodologies~\cite{16,xu2022likert,xia2023skating,du2023learning,CoFInAl,tip/ZengZ24}, we train separate models to predict each score category.

\noindent\textbf{Rhythmic Gymnastics (RG).} The RG dataset contains 1,000 videos of rhythmic gymnastics performances across four apparatus types: ball, clubs, hoop, and ribbon. Each video is approximately 1.6 minutes long and recorded at 25 frames per second. The dataset follows a 4:1 training-evaluation split, with 200 training videos and 50 evaluation videos per apparatus type. Following established protocols~\cite{zeng2020hybrid,xu2022likert,CoFInAl,tip/ZengZ24}, we train apparatus-specific models using standardized scoring annotations.

\begin{figure*}[!t]
	\centering
	\includegraphics[width=1.0\linewidth]{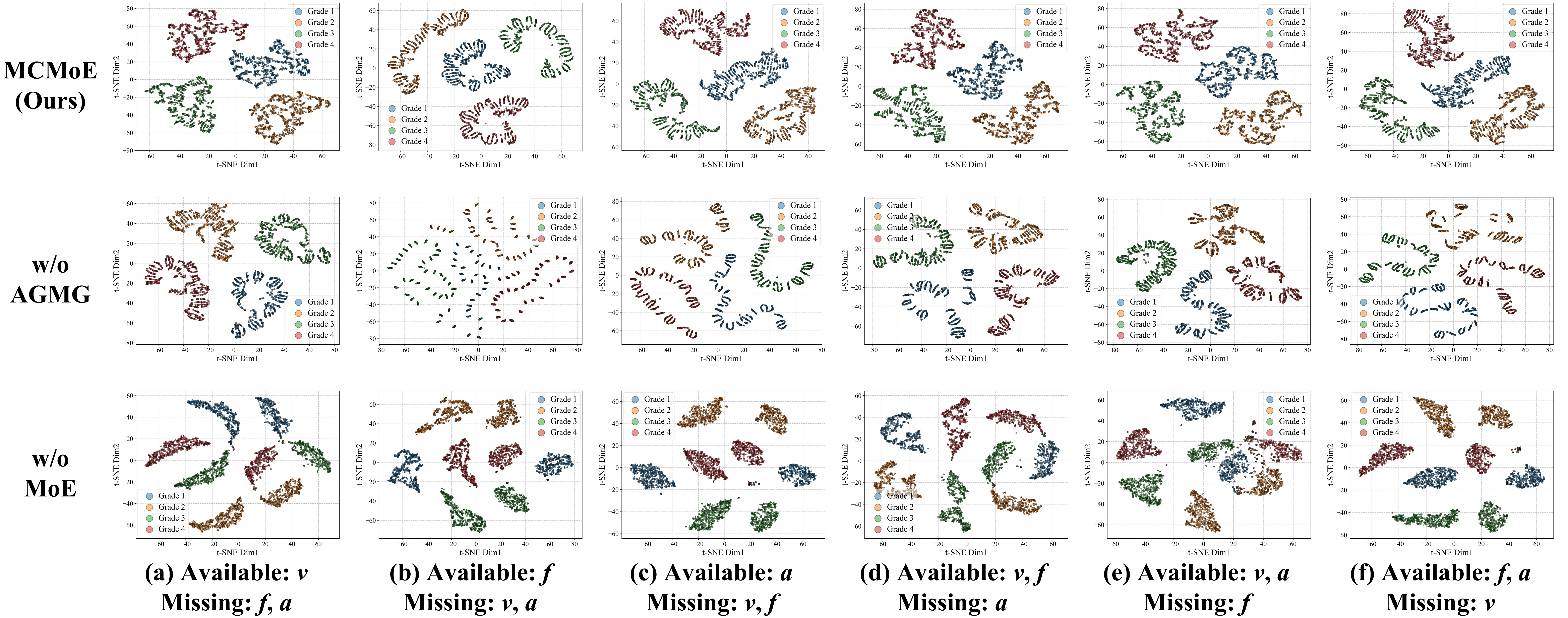}
	\caption{The t-SNE grade feature distribution plots on the FS1000 (PCS) in all incomplete multimodal combinations contrasting with and without our AGMG and MoE. (a), (b), (c), (d), (e) and (f) correspond to incomplete multimodal scenarios for $\{v\}$, $\{f\}$, $\{a\}$, $\{v,f\}$, $\{v,a\}$, and $\{f,a\}$, respectively. (Best viewed in color.)}
	\vspace{-8pt}
	\label{fig:a1}
\end{figure*}
\section{More Ablation Studies}
In this section, we will further conduct some ablation studies to determine the experimental details. Unless otherwise stated, all ablation studies are performed on the RG and Fis-V datasets, containing two types of action scenarios.

\begin{table}[t]
	\centering
	\tabcolsep=2pt
	\resizebox{\linewidth}{!}{
	\begin{tabular}{@{}ccc|cc|cc@{}}
		\toprule
		\multicolumn{3}{c|}{Loss   Settings} & \multicolumn{2}{c|}{RG}       & \multicolumn{2}{c}{Fis-V}     \\ \midrule
		${\mathcal{L}_{recon}}$       & ${{\mathcal L}_{align}}$     & ${{\mathcal L}_{div}}$     & Average       & $\{v,f,a\}$    & Average       & $\{v,f,a\}$     \\ \midrule
		$\checkmark$           &            &          & 0.620 / 9.08  & 0.785 / 5.81 & 0.677 / 18.71 & 0.791 / 15.08 \\
		& $\checkmark$          &          & 0.616 / 10.84 & 0.748 / 7.26 & 0.652 / 18.96 & 0.788 / 15.77 \\
		&            & $\checkmark$        & 0.613 / 12.47 & 0.762 / 6.37 & 0.682 / 20.09 & 0.781 / 14.86 \\
		$\checkmark$           & $\checkmark$          &          & 0.624 / 8.13  & 0.791 / 4.93 & 0.693 / 17.54 & 0.795 / 12.29 \\
		$\checkmark$           &            & $\checkmark$        & 0.648 / 8.56  & 0.797 / 5.16 & 0.708 / 19.25 & 0.790 / 12.59 \\
		& $\checkmark$          & $\checkmark$        & 0.617 / 9.30  & 0.771 / 5.50 & 0.699 / 18.45 & 0.813 / 12.47 \\ 
		$\checkmark$  & $\checkmark$          & $\checkmark$       & \textbf{0.697 / 7.89}  & \textbf{0.842 / 4.85} & \textbf{0.734 / 17.02} & \textbf{0.829} / \textbf{12.15} \\ 
		\bottomrule
	\end{tabular}}
	\caption{Effects of different loss functions. $\checkmark$ represents the use of the loss. Results are shown by Sp. Corr.(↑) / MSE(↓).}
	\label{tab:a1}
\end{table}
\noindent\textbf{Effects of different loss functions.} We supplement the main manuscript with a more comprehensive ablation study of the effects of losses. The basic ${\mathcal{L}_{task}}$ is retained and all combinations of ${\mathcal{L}_{recon}}$, ${{\mathcal L}_{align}}$, and ${{\mathcal L}_{div}}$ are verified. As shown in Table \ref{tab:a1}, all losses have a significant effect, especially for performance in incomplete multimodal scenarios. This shows the rationality and effectiveness of our design of loss functions. Specifically, ${\mathcal{L}_{recon}}$ provides important supervision for missing modality completion, avoiding generating misinformation that can mislead learning. ${{\mathcal L}_{align}}$ motivates the MoE to focus on both unimodal and cross-modal joint representation learning in single-stage training. While ${{\mathcal L}_{div}}$ ensures that each grade pattern focuses on different actions, facilitating accurate AQA.

\begin{table}[t]
	\centering
	\renewcommand\arraystretch{0.95}
	\resizebox{0.9\linewidth}{!}{
	\begin{tabular}{@{}c|cc|cc@{}}
		\toprule
		\multirow{2}{*}{\#$N$} & \multicolumn{2}{c|}{RG}                        & \multicolumn{2}{c}{Fis-V}                       \\ \cmidrule(l{0pt}r{0pt}){2-5} 
		& Average       & $\{v,f,a\}$    & Average       & $\{v,f,a\}$             \\ \midrule
		2                    & 0.673 / 8.51          & 0.821 / 5.44          & 0.720 / 18.56          & 0.811 / 14.67          \\
		3                    & 0.686 / 7.95          & 0.828 / 5.31          & 0.729 / 17.73          & 0.821 / 14.06          \\
		4                    & 0.697 / \textbf{7.89}          & \textbf{0.842 / 4.85} & \textbf{0.734 / 17.02} & \textbf{0.829 / 12.15} \\
		5                    & \textbf{0.700} / 7.93 & 0.838 / 4.88          & 0.730 / 17.18          & 0.823 / 12.43          \\
		6                    & 0.692 / 8.02          & 0.835 / 4.92          & 0.731 / 17.26          & 0.820 / 12.66          \\ \bottomrule
	\end{tabular}}
	\vspace{-5pt}
	\caption{Effects of the number of grade patterns $N$.}
	\vspace{-10pt}
	\label{tab:a2}
\end{table}
\noindent\textbf{Effects of the number of grade patterns $N$.} We follow the priori works \cite{xu2022likert,CoFInAl,10884538,huang2025dual} using state-of-the-art grade-based regression networks. Each grade pattern represents focusing on actions of a certain performance quality. As shown in Table \ref{tab:a2}, the performance improves significantly when N is raised from 2 to 4, indicating that fine-grained modeling is important for AQA. And when N > 4, the performance tends to decrease. This may be due to excessive grade patterns, which makes it difficult f2or neighboring grade patterns to recognize similar information.

\begin{table}[t]
	\centering
	\tabcolsep=1.5pt
	\resizebox{\linewidth}{!}{
	\begin{tabular}{@{}c|cc|cc@{}}
		\toprule
		\multirow{2}{*}{Methods} & \multicolumn{2}{c|}{RG}                        & \multicolumn{2}{c}{Fis-V}                       \\ \cmidrule(l{0pt}r{0pt}){2-5} 
		& Average       & $\{v,f,a\}$    & Average       & $\{v,f,a\}$             \\ \midrule
		Summation                & 0.675 / 7.91          & 0.817 / 5.30          & 0.701 / 18.07          & 0.790 / 13.10          \\
		Weighted Average         & 0.684 / 7.98          & 0.830 / 5.08          & 0.724 / 17.61          & 0.811 / 12.76          \\
		Full-connection Layer    & 0.690 / 7.90          & 0.834 / 5.00          & 0.728 / 17.19          & 0.819 / 12.66          \\
		\textbf{CFM (Ours)}      & \textbf{0.697 / 7.89} & \textbf{0.842 / 4.85} & \textbf{0.734 / 17.02} & \textbf{0.829 / 12.15} \\ \bottomrule
	\end{tabular}}
	\vspace{-5pt}
	\caption{Effects of the cross-modal fusion module.}
	\vspace{-8pt}
	\label{tab:a6}
\end{table}
\begin{figure}[!t]
	\centering
	\includegraphics[width=\linewidth]{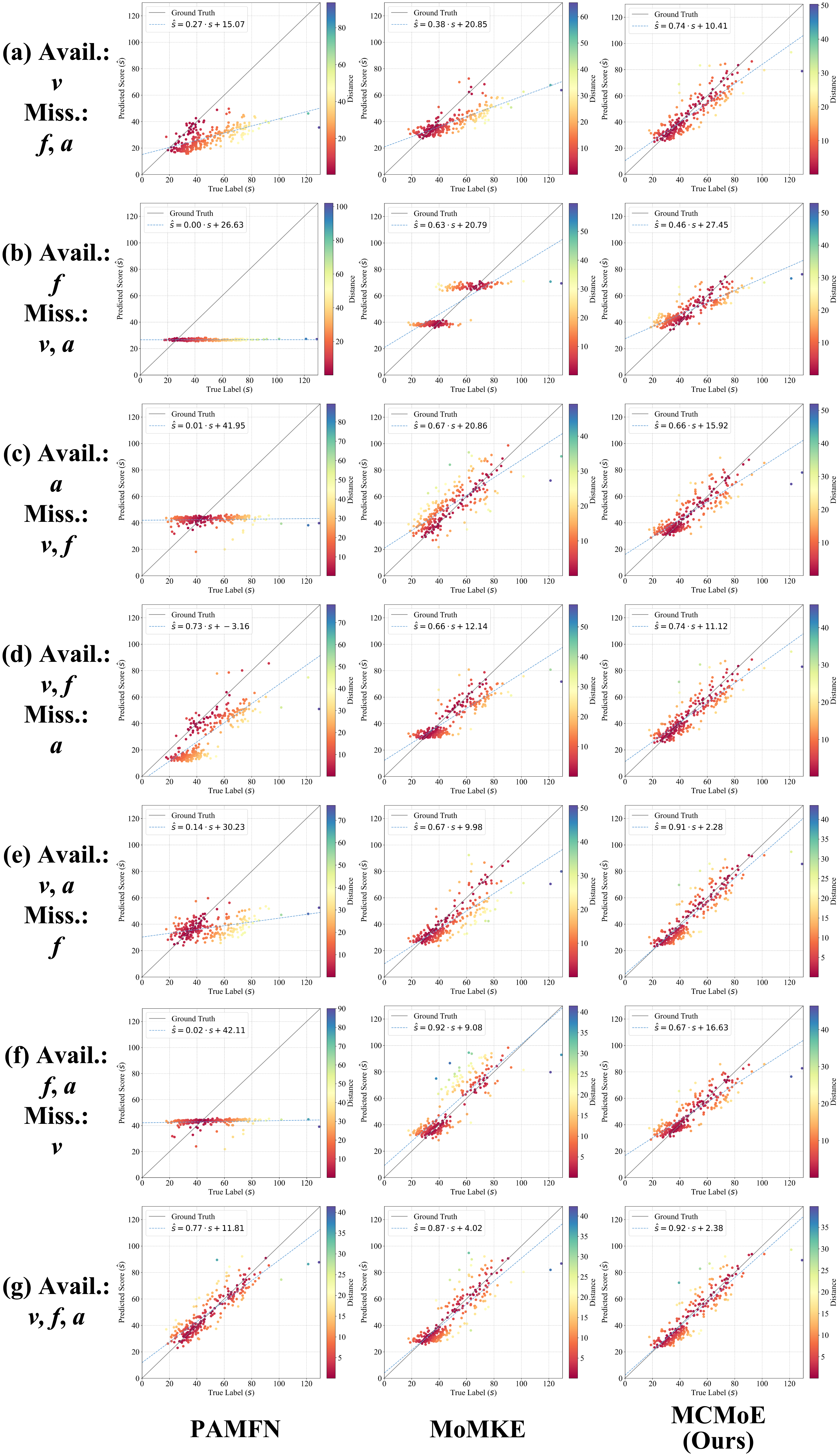}
	\caption{Comparison of scatter plots with the state-of-the-art multimodal AQA method PAMFN~\cite{tip/ZengZ24} and the incomplete multimodal learning method MoMKE~\cite{xu2024leveraging} for all modal combinations on FS1000 (TES). The horizontal axis of each plot represents the true score and the vertical axis is the predicted score. The color of the scatter represents the degree of difference between the predicted and true values.}
	\vspace{-10pt}
	\label{fig:a2}
\end{figure}
\noindent\textbf{Effects of the cross-modal fusion module.} We design a \textbf{C}ross-modal \textbf{F}usion \textbf{M}odule (\textbf{CFM}), implemented as a two-layer convolutional block, to capture inter-modal correlations. Since the features processed by the MoE contain cross-modal joint semantics, our CFM can rely on the simple architecture to achieve effective multimodal fusion. As shown in Table \ref{tab:a6}, we compare our CFM against common fusion strategies like summation, weighted average, and fully-connected layers. The results demonstrate that our CFM significantly improves performance. This may be because the poorly intra-class discriminative AQA task requires a high degree of fine-grained discriminative ability from the model. This challenge is exacerbated by complex semantic learning for incomplete multimodal scenarios, resulting in simple fusion strategies that cannot effectively map multimodal semantics to the quality-aware semantic space.

\section{More Visualizations}
In this section, we provide more visualizations to illustrate the contribution of our proposed designs to incomplete multimodal learning and action quality assessment.

\noindent\textbf{Effects of AGMG and MoE.} To visualize the impact of the complementarity between MMC and MoE that we emphasize, Figure \ref{fig:a1} presents the t-SNE distributions of grade patterns on FS1000 (PCS) after removing the AGMG and MoE components. Each point is a grade feature. Our MCMoE effectively discriminates between varying action qualities, producing compact clusters and clearly defined inter-grade boundaries. In contrast, omitting AGMG or MoE leads to more dispersed distributions and blurred class distinctions. Without AGMG's missing modality completion, MoE struggles to capture intra-class modality consistency, resulting in dispersed intra-grade features. Meanwhile, the absence of MoE's dynamic cross-modal fusion prevents effective collaboration of unimodal semantics, splitting quality grades into two clusters due to modality divergence. This visualization underscores the significant contributions of MMC and MoE to quality space modeling.

\noindent\textbf{Visualization of predicted scores.} To visually compare the quality assessment performance of our method with existing methods, we visualize scatter plots of the prediction scores in Figure \ref{fig:a2}. Each point represents the prediction score of one video sample. It can be seen that our method has significantly better AQA performance. Our MCMoE predicts more accurate scores with tighter correlations in both complete and incomplete multimodal scenarios.

\begin{table*}[t]
	\centering
	\tabcolsep=1.5pt
	\renewcommand\arraystretch{0.9}
	\resizebox{\linewidth}{!}{
		\begin{tabular}{@{}c|c|c|cccccccc@{}}
			\toprule
			\multirow{2}{*}{Datasets} & \multirow{2}{*}{Methods} & \multirow{2}{*}{Year} & \multicolumn{8}{c}{Assessment Category (Spearman Correlation (↑) / Mean Square Error (↓))}                                                                                \\ \cmidrule(l{0pt}r{0pt}){4-11} 
			&                          &                       & TES              & PCS              & SS               & TR               & PE               & CO               & IN               & Average          \\ \midrule
			\multirow{7}{*}{FS1000}   & $\heartsuit$MLP-Mixer               & 2023                  & 0.880 / 81.24    & 0.820 / 9.47     & 0.800 / 0.35     & 0.810 / 0.35     & 0.800 / 0.62     & 0.810 / 0.37     & 0.810 / 0.39     & 0.821 / 13.26    \\
            & $\diamondsuit$T²CR*                    & 2024                  & 0.863 / 107.59   & 0.794 / 15.26    & 0.833 / 0.61     & 0.837 / 0.48     & 0.823 / 0.69     & 0.843 / 0.57     & 0.804 / 0.42     & 0.829 / 17.95    \\
			& $\diamondsuit$CoFInAl*                 & 2024                  & 0.835 / 81.65    & 0.830 / 16.05    & 0.838 / 0.56     & 0.836 / 0.63     & 0.814 / 0.71     & 0.829 / 0.41     & 0.819 / 0.54     & 0.829 / 14.36    \\
			& $\heartsuit$SGN                    & 2024                  & \textbf{0.890} / 79.08    & 0.850 / \textbf{8.40}     & 0.840 / \textbf{0.31}     & 0.850 / \textbf{0.32}     & 0.820 / \textbf{0.61}     & \textbf{0.850} / \textbf{0.33}     & 0.830 / \textbf{0.37}     & 0.849 / \textbf{12.77}    \\
			& $\heartsuit$PAMFN*                  & 2024                  & 0.874 / \textbf{78.42}    & \textbf{0.854} / 9.72     & 0.848 / 0.54     & \textbf{0.866} / 0.58     & \textbf{0.857} / 0.69     & 0.834 / 0.53     & \textbf{0.846} / 0.64     & \textbf{0.855} / 13.02    \\
            & $\diamondsuit$ASGTN*         & 2025                 & 0.873 / 94.83    & 0.849 / 14.58     & \textbf{0.858} / 0.35     & 0.856 / 0.38   & 0.832 / 0.68  & 0.846 / 0.39     & 0.835 / 0.47     & 0.850 / 15.95         \\
			& \textbf{MCMoE(Ours)}             & -                     & {\color{red} \textbf{0.900}} / {\color{red} \textbf{71.69}}    & {\color{red} \textbf{0.872}} / {\color{red} \textbf{7.43}}     & {\color{red} \textbf{0.876}} / {\color{red} \textbf{0.26}}     & {\color{red} \textbf{0.882}} / {\color{red} \textbf{0.26}}     & {\color{red} \textbf{0.872}} / {\color{red} \textbf{0.52}}     & {\color{red} \textbf{0.886}} / {\color{red} \textbf{0.26}}     & {\color{red} \textbf{0.874}} / {\color{red} \textbf{0.29}}     & {\color{red} \textbf{0.881}} / {\color{red} \textbf{11.53}}    \\
			& ${\Delta _{SOTA}}$                     & -                     & {\color{blue} ↑1.1\%} / {\color{blue} ↓8.6\%}  & {\color{blue} ↑2.1\%} / {\color{blue} ↓11.5\%} & {\color{blue} ↑2.1\%} / {\color{blue} ↓16.1\%} & {\color{blue} ↑1.8\%} / {\color{blue} ↓18.8\%} & {\color{blue} ↑1.8\%} / {\color{blue} ↓14.8\%} & {\color{blue} ↑4.2\%} / {\color{blue} ↓21.2\%} & {\color{blue} ↑3.3\%} / {\color{blue} ↓21.6\%} & {\color{blue} ↑3.0\%} / {\color{blue} ↓9.7\%}  \\ \midrule
			\multirow{8}{*}{Fis-V}   		& $\heartsuit$MLP-Mixer               & 2023                  & 0.680 / 19.57    & 0.820 / 7.96     & †                & †                & †                & †                & †                & 0.759 / 13.77    \\
            & $\diamondsuit$T²CR*                    & 2024                  & 0.702 / 20.84    & 0.811 / 8.10     & †                & †                & †                & †                & †                & 0.762 / 14.47    \\
			& $\diamondsuit$CoFInAl                & 2024                  & 0.716 / 20.76    & 0.843 / \textbf{7.91}     & †                & †                & †                & †                & †                & 0.788 / 14.34    \\
			& $\heartsuit$SGN                      & 2024                  & 0.700 / \textbf{19.05}    & 0.830 / 7.96     & †                & †                & †                & †                & †                & 0.773 / \textbf{13.51}    \\
			& $\heartsuit$PAMFN                   & 2024                  & \textbf{0.754} / 22.50    & \textbf{0.872} / 8.16     & †                & †                & †                & †                & †                & \textbf{0.822} / 15.33    \\
            & $\heartsuit$VATP-Net            & 2025                  & 0.702 / -        & 0.863 / -        & †                & †                & †                & †                & †                & 0.796 / -        \\
            & $\diamondsuit$ASGTN*         & 2025       & 0.703 / 21.37    & 0.845 / 10.75     & †                & †                & †                & †                & †                & 0.784 / 16.06    \\
			& \textbf{MCMoE(Ours)}             & -                     & {\color{red} \textbf{0.759}} / {\color{red} \textbf{18.50}}    & {\color{red} \textbf{0.880}} / {\color{red} \textbf{5.81}}     & †                & †                & †                & †                & †                & {\color{red} \textbf{0.829}} / {\color{red} \textbf{12.15}}    \\
			& ${\Delta _{SOTA}}$                     & -                     & {\color{blue} ↑0.7\%} / {\color{blue} ↓2.9\%}  & {\color{blue} ↑0.9\%} / {\color{blue} ↓26.5\%} & †                & †                & †                & †                & †                & {\color{blue} ↑0.9\%} / {\color{blue} ↓10.1\%} \\ \midrule
			\multicolumn{1}{c}{} &  \multicolumn{1}{c}{}                       &      \multicolumn{1}{c}{}                 & Ball             & Clubs            & Hoop             & Ribbon           & †                & †                & †                & Average          \\ \midrule
			\multirow{7}{*}{RG}      & $\heartsuit$MLP-Mixer*              & 2023                  & 0.677 / 6.75     & 0.708 / 5.81     & 0.778 / 5.94     & 0.706 / 6.87     & †                & †                & †                & 0.719 / 6.34     \\
            & $\diamondsuit$T²CR*                    & 2024                  & 0.750 / 7.18     & 0.800 / 5.63     & 0.794 / 5.88     & 0.827 / \textbf{5.81}     & †                & †                & †                & 0.794 / 6.13     \\
            & $\diamondsuit$CoFInAl                 & 2024                  & {\color{red} \textbf{0.809}} / {\color{red} \textbf{5.07}}     & 0.806 / \textbf{5.19}     & 0.804 / 6.37     & 0.810 / 6.30     & †                & †                & †                & 0.807 / \textbf{5.73}     \\
			& $\heartsuit$PAMFN                   & 2024                  & 0.757 / 6.24     & {\color{red} \textbf{0.825}} / 7.45     & \textbf{0.836} / {\color{red} \textbf{5.21}}     & \textbf{0.846} / 7.67     & †                & †                & †                & \textbf{0.819} / 6.64     \\
            & $\heartsuit$VATP-Net                 & 2025                  & 0.800 / -        & 0.810 / -        & 0.780 / -        & 0.769 / -        & †                & †                & †                & 0.790 / -        \\
            & $\diamondsuit$ASGTN*         & 2025                & 0.792 / 6.60     & {\color{red} \textbf{0.825}} / 5.66     & 0.784 / {\color{red} \textbf{5.21}}     & 0.793 / 6.75     & †                & †                & †                & 0.799 / 6.06     \\
			& \textbf{MCMoE(Ours)}             & -                     & \textbf{0.806} / \textbf{5.66}     & \textbf{0.815} / {\color{red} \textbf{4.22}}     & {\color{red} \textbf{0.845}} / \textbf{5.62}     & {\color{red} \textbf{0.890}} / {\color{red} \textbf{3.89}}     & †                & †                & †                & {\color{red} \textbf{0.842}} / {\color{red} \textbf{4.85}}     \\
			& ${\Delta _{SOTA}}$                     & -                     & {\color{gray} ↓0.4\%} / {\color{gray} ↑11.6\%} & {\color{gray} ↓1.2\%} / {\color{blue} ↓18.7\%} & {\color{blue} ↑1.1\%} / {\color{gray} ↑7.9\%}  & {\color{blue} ↑5.2\%} / {\color{blue} ↓33.0\%} & †                & †                & †                & {\color{blue} ↑2.8\%} / {\color{blue} ↓15.4\%} \\ \bottomrule
	\end{tabular}}
	\vspace{-5pt}
    \caption{Comparisons of performance with the state-of-the-art (SOTA) on three benchmarks with complete modalities. The {\color{red} \textbf{red bold}} / \textbf{black bold} indicate the best / second-best results. † means the dataset does not include this category. * indicates our reimplementation based on the official code. ${\Delta _{SOTA}}$ means the performance {\color{blue} increase} or {\color{gray} decrease} of our MCMoE compared to the best competing methods. $\diamondsuit$ and $\heartsuit$ represent unimodal and multimodal AQA methods.}
	\label{tab:2}
	\vspace{-10pt}
\end{table*}
\section{More Experiment Settings}
\label{sec:settings}
\noindent \textbf{Metrics.} We adopt the widely used metrics in AQA, including Spearman's Rank Correlation Coefficient (SRCC/${\rho}$) and Mean Square Error (MSE). MSE measure the numerical difference between the ground-truth score ${y}$ and the predicted score ${\hat y}$. While ${\rho}$ measure the difference between the ground-truth series ${\hat q}$ and the predicted series ${q}$:
\begin{equation}
	\small
	{\text{MSE}}\left( {y,\hat y} \right) = {\left\| {y - \hat y} \right\|^2},
\end{equation}
\begin{equation}
	\small
	\rho  = \frac{{\sum\nolimits_i {\left( {{q_i} - \bar q} \right)\left( {{{\hat q}_i} - \bar{\hat{q}}} \right)} }}{{\sqrt {\sum\nolimits_i {{{\left( {{q_i} - \bar{q}} \right)}^2}\sum\nolimits_i {{{\left( {{{\hat q}_i} - \bar{\hat{q}}} \right)}^2}} } } }}.
\end{equation}

\noindent \textbf{Compute resources.} All experiments are performed on one RTX 3090 GPU equipped with PyTorch 1.12.0 and a CPU at 2.40GHz. The version of CUDA is 11.3. For example, it takes about four and a half hours to train the RG dataset with a batch size of 32 and 500 epochs using visual, flow, and audio features extracted from the pre-trained backbone.

\noindent\textbf{Label normalization.} Existing datasets often have different ranges of score labels, increasing the challenge of robustness of the regression network. To address this, following established methodologies~\cite{zeng2020hybrid, xu2022likert, tip/ZengZ24, CoFInAl}, we normalize score labels to a unified [0,1] range using a scaling factor $\xi$. Specifically, for real-valued score labels ${{s}_{i}}$ in a dataset, normalized labels ${\hat s}_{i}$ are computed as ${{s}_{i}}/\xi$, where $\xi$ is determined by the maximum score in the training set. In our experiments, $\xi$ is set to 130, 60, and 10 for FS1000's TES, PCS, and the remaining sub-classes (SS, TR, PE, CO, IN), respectively. For Fis-V(TES), Fis-V(PCS), and RG, $\xi$ is set to 45, 40, and 25, respectively. To ensure fair comparisons with existing methods, our predicted scores are scaled back to their original ranges by multiplying with $\xi$ when computing the MSE metric.

\noindent\textbf{Learning strategies and epoch settings.} We implement a cosine annealing strategy to dynamically adjust the learning rate during training, following established practices. To optimize convergence, we adopt dataset-specific epoch settings across models, as done in prior works~\cite{zeng2020hybrid, xu2022likert, tip/ZengZ24, CoFInAl}. For the FS1000 dataset, epoch settings are as follows: TES (360), PCS (460), SS (360), TR (210), PE (520), CO (520), and IN (390). For Fis-V, the epochs are set to 460 for TES and 510 for PCS. The RG dataset uses 410 epochs for Ball, 560 for Clubs, 270 for Hoop, and 300 for Ribbon. These settings ensure effective model training and alignment with the unique characteristics of each dataset.

\noindent\textbf{Implementation details of the architecture.} Here, we give the specific details of each component in the proposed MCMoE. Our shared temporal enhancement module (STEM) is implemented by a three-layer Transformer encoder with two-head self-attention. The proposed adaptive gated modality generator (AGMG) is implemented by two-layer four-head cross-attention and one gating layer. The two-layer MLP constituting the unimodal expert contains 3×3 convolution, batch normalization and GELU nonlinearity. The two-layer MLP implementing the soft router is fully connected layers with GELU nonlinearity.

\section{More Comparisons of the SOTA Methods}
In the main manuscript, extensive experiments show that our approach achieves state-of-the-art performance on three long-term AQA benchmarks in both complete and incomplete multimodal scenarios. In this section, we provide more detailed comparisons with state-of-the-art (SOTA) methods on the three benchmarks.

Firstly, for a more detailed comparison, Table \ref{tab:2} presents the complete results under full-modality settings, which were visualized as bar charts in the main manuscript. We compare our method with SOTA unimodal (T²CR~\cite{ke2024two}, CoFInAl~\cite{CoFInAl}, and ASGTN~\cite{10884538}) and multimodal (MLP-Mixer~\cite{xia2023skating}, SGN~\cite{du2023learning}, VATP-Net~\cite{gedamu2024visual}, and PAMFN~\cite{tip/ZengZ24}) AQA methods. The results show that our MCMoE achieves the best performance on all three datasets, outperforming the second-best method by 2.2\% of SP. Corr. and 11.7\% of MSE on average. This demonstrates the effectiveness of our proposed MCMoE in multimodal AQA.

Secondly, due to space constraints, we mainly show the average metrics for all action/score categories under a given experimental setting in the main manuscript. Here, we present the complete comparisons with existing state-of-the-art methods on incomplete multimodal AQA on the three datasets. The experimental results in Tables \ref{tab:a3}, \ref{tab:a4}, and \ref{tab:a5} correspond to the FS1000, Fis-V, and RG datasets, respectively.

\emph{\textbf{Please refer to the experimental results on the page following the references.}}

\begin{table*}[t]
	\centering
	\tabcolsep=1.5pt
	\resizebox{\linewidth}{!}{
	\begin{tabular}{@{}c|c|cccccccc@{}}
		\toprule
		&        & \multicolumn{8}{c}{Testing Condition (Spearman Correlation (↑) / Mean Square Error (↓))}      \\ \cmidrule(l{0pt}r{0pt}){3-10}
		\multirow{-2}{*}{Types} & \multirow{-2}{*}{Methods}& $\{v,f\}$          & $\{v,a\}$           & $\{f,a\}$           & $\{v\}$             & $\{f\}$            & $\{a\}$             & Average           & $\{v,f,a\}$                              \\ \midrule
		& $\heartsuit$MLP-Mixer*                 & 0.749 / 156.74                                & 0.366 / 380.63                                & 0.015 / 573.44                                & 0.286 / 675.68                                & 0.276 / 578.91                                 & 0.457 / 429.42                                & 0.386 / 465.80                                & 0.877 / 86.54                                 \\
		& $\heartsuit$PAMFN*                     & 0.827 / 388.21                                & 0.351 / 397.58                                & 0.434 / 360.58                                & 0.677 / 607.79                                & 0.011 / 788.74                                 & 0.241 / 372.50                                & 0.474 / 485.90                                & 0.874 / \textbf{78.42}                                 \\
		& $\clubsuit$ActionMAE*                 & 0.853 / 144.66                       & 0.833 / 410.45                                & \textbf{0.862} / 138.01                                & 0.800 / 328.56                                & 0.778 / {\color{red} \textbf{118.02}}                                 & 0.762 / 256.94                                & 0.818 / 232.77                                & \textbf{0.881} / 107.07                       \\
		& $\spadesuit$GCNet*                     & 0.813 / 152.23                                & 0.776 / \textbf{142.94}                                & 0.801 / 138.07                                & 0.763 / \textbf{152.35}                                & 0.785 / 183.14                                 & 0.741 / 233.26                                & 0.781 / 167.00                                & 0.820 / 129.12                                \\
		& $\spadesuit$IMDer*                    & 0.847 / 122.66                                & 0.827 / 179.14                                & 0.833 / 137.80                                & 0.737 / 213.77                                & 0.742 / \textbf{123.89}                                 & \textbf{0.805} / 194.01                       & 0.802 / 161.88                                & 0.853 / 159.36                                \\
		& $\spadesuit$MoMKE*                 & 0.843 / 108.53                                & \textbf{0.853} / 146.50                       & 0.854 / 141.05                       & \textbf{0.813} / 234.39                       & 0.788 / 126.05                        & 0.798 / \textbf{158.86}                                & \textbf{0.827} / 152.56                       & 0.876 / 97.97                                 \\
		& $\spadesuit$SDR-GNN*                   & \textbf{0.861} / \textbf{98.49}                                 & 0.807 / 153.41                                & 0.856 / \textbf{120.34}                                & 0.810 / 164.88                                & \textbf{0.811} / 177.01                                 & 0.778 / 141.63                                & 0.823 / \textbf{142.63}                                & 0.872 / 88.89                                 \\
		& \textbf{MCMoE(Ours)}     & {\color{red} \textbf{0.883 / 77.76}} & {\color{red} \textbf{0.901 / 73.64}} & {\color{red} \textbf{0.878 / 81.65}} & {\color{red} \textbf{0.879 / 84.11}} & {\color{red} \textbf{0.835}} / 129.01 & {\color{red} \textbf{0.858 / 91.90}} & {\color{red} \textbf{0.874 / 89.68}} & {\color{red} \textbf{0.900 / 71.69}} \\
		\multirow{-9}{*}{TES}        & ${\Delta _{SOTA}}$                      & {\color{blue} ↑2.6\% / ↓21.0\%}       & {\color{blue} ↑5.6\% / ↓48.5\%}       & {\color{blue} ↑1.9\% / ↓32.2\%}       & {\color{blue} ↑8.1\% / ↓44.8\%}       & {\color{blue} ↑3.0\%} / {\color{gray} ↑9.3\%}         & {\color{blue} ↑6.6\% / ↓42.2\%}       & {\color{blue} ↑5.7\% / ↓37.1\%}       & {\color{blue} ↑2.2\% / ↓8.6\%}        \\ \midrule
		& $\heartsuit$MLP-Mixer*               & 0.714 / \textbf{17.95}                                 & 0.571 / 32.46                                 & 0.609 / \textbf{26.25}                                 & 0.717 / 24.71                                 & 0.595 / 26.09                                  & 0.005 / 37.12                                 & 0.565 / 27.43                                 & 0.799 / 12.47                                 \\
		& $\heartsuit$PAMFN*                   & 0.720 / 25.49                                 & 0.496 / 26.08                                 & 0.612 / 26.54                                 & 0.722 / 24.39                                 & \textbf{0.618 / 24.98}                         & 0.266 / 42.26                                 & 0.590 / 28.29                                 & \textbf{0.854 / 9.72}                         \\
		& $\clubsuit$ActionMAE*                 & 0.734 / 25.13                                 & 0.530 / 35.87                                 & 0.613 / 42.83                        & \textbf{0.745 / 23.18}                        & 0.440 / 26.44                                  & 0.314 / 28.72                                 & 0.583 / 30.36                                 & 0.784 / 16.04                                 \\
		& $\spadesuit$GCNet*                     & 0.674 / 21.87                                 & 0.691 / 20.68                                 & 0.477 / 28.16                                 & 0.675 / 26.05                                 & 0.476 / 28.68                                  & 0.401 / 34.32                                 & 0.577 / 26.63                                 & 0.692 / 20.61                                 \\
		& $\spadesuit$IMDer*                     & 0.688 / 28.16                                 & 0.602 / \textbf{16.55}                                 & 0.608 / 30.15                                 & 0.594 / 34.39                                 & 0.592 / 29.67                                  & 0.466 / \textbf{27.89}                                 & 0.595 / 27.80                                 & 0.725 / 19.53                                 \\
		& $\spadesuit$MoMKE*                   & 0.784 / 20.65                        & \textbf{0.791} / 17.68                        & 0.528 / 27.83                                 & \textbf{0.745} / 27.95                        & 0.314 / 30.86                                  & \textbf{0.481} / 28.28                        & 0.638 / \textbf{25.54}                        & 0.800 / 17.61                                 \\
		& $\spadesuit$SDR-GNN*                   & \textbf{0.794} / 20.83                                 & 0.773 / 19.07                                 & \textbf{0.614} / 28.77                                 & 0.694 / 30.18                                 & 0.491 / 26.84                                  & 0.479 / 30.43                                 & \textbf{0.658} / 26.02                                 & 0.801 / 19.73                                 \\
		& \textbf{MCMoE(Ours)}     & {\color{red} \textbf{0.836 / 8.92}}  & {\color{red} \textbf{0.866 / 7.72}}  & {\color{red} \textbf{0.699 / 18.99}} & {\color{red} \textbf{0.830 / 9.49}}  & {\color{red} \textbf{0.619 / 24.03}}  & {\color{red} \textbf{0.584 / 20.66}} & {\color{red} \textbf{0.759 / 14.97}} & {\color{red} \textbf{0.872 / 7.43}}  \\
		\multirow{-9}{*}{PCS}        & ${\Delta _{SOTA}}$                      & {\color{blue} ↑5.3\% / ↓50.3\%}       & {\color{blue} ↑9.5\% / ↓53.4\%}       & {\color{blue} ↑13.8\% / ↓27.7\%}      & {\color{blue} ↑11.4\% / ↓59.1\%}      & {\color{blue} ↑0.2\% / ↓3.8\%}         & {\color{blue} ↑21.4\% / ↓25.9\%}      & {\color{blue} ↑15.3\% / ↓41.4\%}      & {\color{blue} ↑2.1\% / ↓23.6\%}       \\ \midrule
		& $\heartsuit$MLP-Mixer*                 & 0.711 / 0.64                                  & 0.313 / 3.59                                  & \textbf{0.678} / 0.88                                  & 0.545 / 2.39                                  & \textbf{0.530} / 0.96                                   & 0.041 / 3.97                                  & 0.498 / 2.07                                  & 0.803 / 0.68                                  \\
		& $\heartsuit$PAMFN*                    & 0.533 / 0.82                                  & 0.664 / 11.80                                 & 0.624 / \textbf{0.74}                                  & 0.746 / 5.45                                  & 0.448 / \textbf{0.91}                                   & 0.040 / 12.31                                 & 0.538 / 5.34                                  & \textbf{0.848} / 0.54                                  \\
		& $\clubsuit$ActionMAE*                & 0.694 / 0.52                                  & 0.750 / \textbf{0.47}                                  & 0.474 / 0.88                                  & 0.689 / 0.55                                  & 0.491 / 1.06                                   & \textbf{0.449 / 0.92}                                  & 0.605 / \textbf{0.73}                                  & 0.788 / 0.46                                  \\
		& $\spadesuit$GCNet*                     & 0.723 / 1.45                                  & 0.577 / 0.87                                  & 0.449 / 1.76                                  & 0.670 / 1.15                                  & 0.304 / 1.11                                   & 0.322 / 1.46                                  & 0.527 / 1.30                                  & 0.727 / 0.61                                  \\
		& $\spadesuit$IMDer*                    & 0.761 / 2.48                                  & 0.640 / 0.82                                  & 0.548 / 2.20                                  & 0.752 / \textbf{0.54}                                  & 0.279 / 2.40                                   & 0.282 / 1.23                                  & 0.575 / 1.61                                  & 0.776 / 0.53                                  \\
		& $\spadesuit$MoMKE*                   & \textbf{0.789} / 0.43                                  & 0.702 / 0.99                                  & 0.392 / 1.87                                  & \textbf{0.767} / 0.97                                  & 0.288 / 1.17                                   & 0.335 / 1.50                                  & 0.584 / 1.16                                  & 0.794 / \textbf{0.40}                                  \\
		& $\spadesuit$SDR-GNN*                   & 0.777 / \textbf{0.40}                                  & \textbf{0.760} / 0.80                                  & 0.418 / 2.07                                  & 0.759 / 0.68                                  & 0.488 / 1.33                                   & 0.308 / 1.74                                  & \textbf{0.617} / 1.17                                  & 0.799 / 0.44                                  \\
		& \textbf{MCMoE(Ours)}     & {\color{red} \textbf{0.844 / 0.32}}  & {\color{red} \textbf{0.876 / 0.27}}  & {\color{red} \textbf{0.687 / 0.63}}  & {\color{red} \textbf{0.841 / 0.31}}  & {\color{red} \textbf{0.563 / 0.76}}   & {\color{red} \textbf{0.541 / 0.82}}  & {\color{red} \textbf{0.755 / 0.52}}  & {\color{red} \textbf{0.876 / 0.26}}  \\
		\multirow{-9}{*}{SS}         & ${\Delta _{SOTA}}$                      & {\color{blue} ↑7.0\% / ↓20.0\%}       & {\color{blue} ↑15.3\% / ↓42.6\%}      & {\color{blue} ↑1.3\% / ↓14.9\%}       & {\color{blue} ↑9.6\% / ↓42.6\%}       & {\color{blue} ↑6.2\% / ↓16.5\%}        & {\color{blue} ↑20.5\% / ↓10.9\%}      & {\color{blue} ↑22.4\% / ↓28.8\%}      & {\color{blue} ↑3.3\% / ↓35.0\%}       \\ \midrule
		& $\heartsuit$MLP-Mixer*                 & 0.689 / 0.89                                  & 0.252 / 1.11                                  & \textbf{0.578} / 1.65                                  & 0.590 / 0.91                                  & 0.555 / 1.19                                   & 0.056 / 1.40                                  & 0.478 / 1.19                                  & 0.805 / \textbf{0.38}                                  \\
		& $\heartsuit$PAMFN*                    & 0.615 / 1.25                                  & 0.565 / 1.08                                  & 0.545 / 1.18                                  & 0.677 / 1.04                                  & \textbf{0.600 / 0.86}                                   & 0.116 / 1.32                                  & 0.537 / 1.12                                  & \textbf{0.866} / 0.58                                  \\
		& $\clubsuit$ActionMAE*                & 0.778 / 0.64                                  & 0.787 / 0.56                                  & 0.456 / 0.94                                  & 0.746 / 0.61                                  & 0.412 / 1.18                                   & 0.442 / 1.07                                  & \textbf{0.632} / 0.83                                  & 0.798 / 0.48                                  \\
		& $\spadesuit$GCNet*                    & 0.661 / 0.92                                  & 0.742 / 0.77                                  & 0.394 / 1.97                                  & 0.691 / 0.80                                  & 0.293 / 1.76                                   & 0.459 / 1.02                                  & 0.562 / 1.21                                  & 0.748 / 0.46                                  \\
		& $\spadesuit$IMDer*                     & 0.759 / 0.76                                  & 0.797 / 0.53                                  & 0.492 / \textbf{0.87}                                  & 0.760 / 0.59                                  & 0.308 / 1.02                                   & \textbf{0.461 / 0.92}                                  & 0.629 / 0.78                                  & 0.792 / 0.44                                  \\
		& $\spadesuit$MoMKE*                  & \textbf{0.779 / 0.45}                                  & \textbf{0.799 / 0.42}                                  & 0.443 / 1.01                                  & \textbf{0.781 / 0.45}                                  & 0.302 / 1.27                                   & 0.413 / 0.99                                  & 0.626 / \textbf{0.77}                                  & 0.795 / 0.41                                  \\
		& $\spadesuit$SDR-GNN*                   & 0.774 / 0.83                                  & 0.758 / 0.67                                  & 0.414 / 1.61                                  & 0.728 / 0.76                                  & 0.323 / 1.08                                   & 0.445 / 0.99                                  & 0.604 / 0.99                                  & 0.807 / 0.57                                  \\
		& \textbf{MCMoE(Ours)}     & {\color{red} \textbf{0.841 / 0.33}}  & {\color{red} \textbf{0.891 / 0.26}}  & {\color{red} \textbf{0.702 / 0.56}}  & {\color{red} \textbf{0.850 / 0.31}}  & {\color{red} \textbf{0.629 / 0.70}}   & {\color{red} \textbf{0.513 / 0.85}}  & {\color{red} \textbf{0.767 / 0.50}}  & {\color{red} \textbf{0.882 / 0.26}}  \\
		\multirow{-9}{*}{TR}         & ${\Delta _{SOTA}}$                      & {\color{blue} ↑8.0\% / ↓26.7\%}       & {\color{blue} ↑11.5\% / ↓38.1\%}      & {\color{blue} ↑21.5\% / ↓35.6\%}      & {\color{blue} ↑8.8\% / ↓31.1\%}       & {\color{blue} ↑4.8\% / ↓18.6\%}        & {\color{blue} ↑11.3\% / ↓7.6\%}       & {\color{blue} ↑21.4\% / ↓35.1\%}      & {\color{blue} ↑1.8\% / ↓31.6\%}       \\ \midrule
		& $\heartsuit$MLP-Mixer*                 & 0.687 / 1.02                                  & 0.677 / 3.13                                  & 0.396 / 1.94                                  & 0.468 / 4.09                                  & 0.459 / 1.84                                   & 0.217 / 1.87                                  & 0.502 / 2.31                                  & 0.838 / 0.79                                  \\
		& $\heartsuit$PAMFN*                     & 0.780 / 1.26                                  & 0.742 / 1.02                                  & \textbf{0.524} / 3.00                                  & 0.764 / 1.08                                  & \textbf{0.496} / 3.16                                   & 0.034 / 1.61                                  & 0.601 / 1.85                                  & \textbf{0.857} / 0.69                                  \\
		& $\clubsuit$ActionMAE*                 & 0.784 / \textbf{0.64}                                  & 0.781 / 0.72                                  & 0.378 / 1.19                                  & \textbf{0.783 / 0.65}                                  & 0.292 / 1.34                                   & 0.255 / 2.02                                  & 0.595 / 1.09                                  & 0.790 / 0.72                                  \\
		& $\spadesuit$GCNet*                    & 0.774 / 0.66                                  & 0.796 / 0.68                                  & 0.406 / 1.17                                  & 0.752 / 1.67                                  & 0.388 / \textbf{1.12}                                   & 0.367 / 1.24                                  & 0.617 / 1.09                                  & 0.793 / 0.83                                  \\
		& $\spadesuit$IMDer*                     & 0.743 / 1.03                                  & 0.803 / \textbf{0.62}                                  & 0.459 / 1.16                                  & 0.764 / 0.72                                  & 0.306 / 1.37                                   & \textbf{0.427} / 1.15                                  & 0.619 / 1.01                                  & 0.800 / 0.75                                  \\
		& $\spadesuit$MoMKE*                  & 0.774 / 0.79                                  & \textbf{0.809 / 0.62}                                  & 0.451 / \textbf{1.09}                                  & 0.777 / 0.74                                  & 0.282 / 1.35                                   & 0.416 / \textbf{1.14}                                  & 0.626 / \textbf{0.95}                                  & 0.803 / \textbf{0.67}                                  \\
		& $\spadesuit$SDR-GNN*                   & \textbf{0.805} / 0.67                                  & 0.794 / 0.69                                  & 0.428 / 1.13                                  & 0.775 / 1.51                                  & 0.403 / 1.19                                   & 0.374 / 1.20                                  & \textbf{0.636} / 1.06                                  & 0.817 / 0.79                                  \\
		& \textbf{MCMoE(Ours)}     & {\color{red} \textbf{0.828 / 0.59}}  & {\color{red} \textbf{0.871 / 0.52}}  & {\color{red} \textbf{0.703 / 0.94}}  & {\color{red} \textbf{0.830 / 0.59}}  & {\color{red} \textbf{0.616 / 1.01}}   & {\color{red} \textbf{0.536 / 1.13}}  & {\color{red} \textbf{0.754 / 0.80}}  & {\color{red} \textbf{0.872 / 0.52}}  \\
		\multirow{-9}{*}{PE}         & ${\Delta _{SOTA}}$                      & {\color{blue} ↑2.9\% / ↓7.8\%}        & {\color{blue} ↑7.7\% / ↓16.1\%}       & {\color{blue} ↑34.2\% / ↓13.8\%}      & {\color{blue} ↑6.0\% / ↓9.2\%}        & {\color{blue} ↑24.2\% / ↓9.8\%}        & {\color{blue} ↑25.5\% / ↓0.9\%}       & {\color{blue} ↑18.6\% / ↓15.8\%}      & {\color{blue} ↑1.8\% / ↓22.4\%}       \\ \midrule
		& $\heartsuit$MLP-Mixer*                & 0.732 / 0.72                                  & 0.660 / 1.96                                  & 0.452 / 2.93                                  & 0.779 / 1.03                                  & \textbf{0.511} / 1.24                                   & 0.153 / 2.18                                  & 0.580 / 1.68                                  & 0.784 / 0.36                                  \\
		& $\heartsuit$PAMFN*                     & 0.802 / 0.57                                  & 0.763 / 1.42                                  & \textbf{0.606} / 1.68                                  & 0.773 / 1.39                                  & 0.506 / 1.81                                   & 0.233 / 1.40                                  & 0.648 / 1.38                                  & 0.834 / 0.53                                  \\
		& $\clubsuit$ActionMAE*                 & 0.768 / \textbf{0.45}                                  & 0.805 / 0.44                                  & 0.442 / \textbf{0.88}                                  & 0.770 / \textbf{0.46 }                                 & 0.397 / \textbf{1.13}                                   & 0.412 / 1.03                                  & 0.633 / \textbf{0.73}                                  & 0.803 / 0.49                                  \\
		& $\spadesuit$GCNet*                    & 0.729 / 0.47                                  & 0.775 / 0.48                                  & 0.486 / 1.25                                  & 0.729 / 0.49                                  & 0.279 / 1.50                                   & 0.361 / 1.64                                  & 0.592 / 0.97                                  & 0.764 / 0.58                                  \\
		& $\spadesuit$IMDer*                     & 0.745 / 0.58                                  & 0.782 / 0.43                                  & 0.440 / 0.91                                  & 0.753 / 0.67                                  & 0.288 / 1.14                                   & 0.408 / 0.95                                  & 0.604 / 0.78                                  & 0.774 / 0.45                                  \\
		& $\spadesuit$MoMKE*                   & \textbf{0.821} / 0.48                                  & \textbf{0.841} / 0.45                                  & 0.439 / 0.93                                  & \textbf{0.819} / 0.54                                  & 0.312 / 1.17                                   & \textbf{0.433} / 0.95                                  & \textbf{0.664} / 0.75                                  & \textbf{0.841 / 0.39}                                  \\
		& $\spadesuit$SDR-GNN*                   & 0.758 / \textbf{0.45}                                  & 0.798 / \textbf{0.35}                                  & 0.515 / 1.03                                  & 0.769 / 0.51                                  & 0.401 / 1.21                                   & 0.389 / \textbf{0.93}                                  & 0.636 / 0.75                                  & 0.789 / 0.55                                  \\
		& \textbf{MCMoE(Ours)}     & {\color{red} \textbf{0.850 / 0.31}}  & {\color{red} \textbf{0.887 / 0.26}}  & {\color{red} \textbf{0.710 / 0.81}}  & {\color{red} \textbf{0.848 / 0.34}}  & {\color{red} \textbf{0.610 / 1.03}}   & {\color{red} \textbf{0.575 / 0.85}}  & {\color{red} \textbf{0.773 / 0.60}}  & {\color{red} \textbf{0.886 / 0.26}}  \\ 
		\multirow{-9}{*}{CO}         & ${\Delta _{SOTA}}$                      & {\color{blue} ↑3.5\% / ↓31.1\%}       & {\color{blue} ↑5.5\% / ↓25.7\%}       & {\color{blue} ↑17.2\% / ↓8.0\%}       & {\color{blue} ↑3.5\% / ↓26.1\%}       & {\color{blue} ↑19.4\% / ↓8.8\%}        & {\color{blue} ↑32.8\% / ↓8.6\%}      & {\color{blue} ↑16.4\% / ↓17.8\%}      & {\color{blue} ↑5.4\% / ↓27.8\%}       \\ \midrule
		& $\heartsuit$MLP-Mixer*                & 0.766 / 0.96                                  & 0.769 / 1.12                                  & 0.459 / 3.31                                  & \textbf{0.790} / 0.65                                  & 0.332 / 2.88                                   & 0.272 / 3.02                                  & 0.606 / 1.99                                  & 0.809 / 0.70                                  \\
		& $\heartsuit$PAMFN*                     & 0.721 / 2.09                                  & 0.780 / 0.65                                  & \textbf{0.561 / 0.80}                                  & 0.600 / 3.57                                  & {\color{red} \textbf{0.606}} / 2.97   & 0.077 / 3.51                                  & 0.590 / 2.27                                  & \textbf{0.846} / 0.64                                  \\
		& $\clubsuit$ActionMAE*                 & 0.783 / \textbf{0.57}                                  & 0.795 / \textbf{0.43}                                  & 0.461 / 0.87                                  & 0.779 / \textbf{0.49}                                  & 0.286 / 1.13                                   & 0.446 / \textbf{0.92}                                  & \textbf{0.630 / 0.73}                                  & 0.798 / 0.45                                  \\
		& $\spadesuit$GCNet*                     & 0.709 / 1.33                                  & 0.777 / 0.59                                  & 0.408 / 2.44                                  & 0.557 / 4.18                                  & 0.442 / 1.57                                   & 0.337 / 2.85                                  & 0.561 / 2.16                                  & 0.787 / 0.53                                  \\
		& $\spadesuit$IMDer*                     & 0.751 / 0.74                                  & 0.690 / 1.10                                  & 0.493 / 0.93                                  & 0.674 / 1.28                                  & 0.304 / \textbf{0.96}                                   & 0.414 / 1.76                                  & 0.576 / 1.13                                  & 0.776 / 0.60                                  \\
		& $\spadesuit$MoMKE*                   & \textbf{0.784} / 0.66                                  & \textbf{0.810} / 0.53                                  & 0.485 / 0.96                                  & 0.783 / 0.67                                  & 0.327 / 1.28                                   & \textbf{0.479} / 0.97                                  & 0.648 / 0.84                                  & 0.811 / 0.51                                  \\
		& $\spadesuit$SDR-GNN*                   & 0.732 / 0.86                                  & 0.800 / 0.57                                  & 0.508 / 1.09                                  & 0.685 / 0.79                                  & 0.458 / 1.08                                   & 0.440 / 1.27                                  & 0.625 / 0.94                                  & 0.822 / \textbf{0.43}                                  \\
		& \textbf{MCMoE(Ours)}     & {\color{red} \textbf{0.823 / 0.37}}  & {\color{red} \textbf{0.877 / 0.29}}  & {\color{red} \textbf{0.724 / 0.61}}  & {\color{red} \textbf{0.831 / 0.35}}  & \textbf{0.602} / {\color{red}\textbf{0.74}}   & {\color{red} \textbf{0.562 / 0.80}}  & {\color{red} \textbf{0.759 / 0.53}}  & {\color{red} \textbf{0.874 / 0.29}}  \\
		\multirow{-9}{*}{IN}         & ${\Delta _{SOTA}}$                      & {\color{blue} ↑5.0\% / ↓35.1\%}       & {\color{blue} ↑8.3\% / ↓32.6\%}       & {\color{blue} ↑29.1\% / ↓23.8\%}      & {\color{blue} ↑5.2\% / ↓28.6\%}       & {\color{gray}  ↓0.7\%} / {\color{blue} ↓22.9\%}        & {\color{blue} ↑17.3\% / ↓13.0\%}      & {\color{blue} ↑17.1\% / ↓27.4\%}      & {\color{blue} ↑3.3\% / ↓32.6\%}       \\ \bottomrule
	\end{tabular}}
    \caption{omparisons with the state-of-the-art on the FS1000 under incomplete multimodal scenarios. $v$, $f$, and $a$ refer to the RGB, flow, and audio modalities. ``Average'' denotes the average result of all six incomplete multimodal combinations. The {\color{red} \textbf{red bold}} / \textbf{black bold} indicate the best / second-best results. * indicates our reimplementation. ${\Delta _{SOTA}}$ means the performance {\color{blue} increase} or {\color{gray} decrease} of our MCMoE compared to the best competing methods. $\heartsuit$, $\clubsuit$, and $\spadesuit$ mean the evaluated method sources for multimodal AQA, incomplete multimodal action recognition, and incomplete multimodal emotion recognition.}
	\label{tab:a3}
\end{table*}

\begin{table*}[t]
	\centering
	\tabcolsep=1.5pt
	\resizebox{\linewidth}{!}{
	\begin{tabular}{@{}c|c|cccccccc@{}}
		\toprule
		&        & \multicolumn{8}{c}{Testing Condition (Spearman Correlation (↑) / Mean Square Error (↓))}      \\ \cmidrule(l{0pt}r{0pt}){3-10}
		\multirow{-2}{*}{Types} & \multirow{-2}{*}{Methods}& $\{v,f\}$          & $\{v,a\}$           & $\{f,a\}$           & $\{v\}$             & $\{f\}$            & $\{a\}$             & Average           & $\{v,f,a\}$                              \\ \midrule
		& $\heartsuit$MLP-Mixer*                & 0.662 / 52.37                                 & 0.547 / 81.73                                 & 0.533 / 38.49                                 & 0.445 / 86.26                                 & 0.471 / 40.38                                  & 0.143 / 113.84                                & 0.480 / 68.85                                 & 0.707 / \textbf{20.06}                                 \\
		& $\heartsuit$PAMFN                     & \textbf{0.717} / 58.53                                 & 0.429 / 79.98                                 & \textbf{0.666} / 169.36                                & 0.382 / 140.72                                & {\color{red} \textbf{0.647}} / 162.33 & 0.202 / 89.21                                 & 0.530 / 116.69                                & \textbf{0.754} / 22.50                                 \\
		& $\clubsuit$ActionMAE*                 & 0.627 / 43.24                                 & \textbf{0.652} / 39.91                                 & 0.482 / 32.07                                 & 0.590 / 61.39                                 & 0.387 / 30.51                                  & \textbf{0.456} / {\color{red} \textbf{37.13}}                                 & 0.539 / 40.71                                 & 0.645 / 23.60                                 \\
		& $\spadesuit$GCNet*                     & 0.678 / 19.67                                 & 0.601 / 28.03                                 & 0.553 / 27.16                                 & 0.601 / 27.20                                 & 0.619 / 24.50                                  & 0.390 / 43.17                                 & 0.580 / \textbf{28.29}                                 & 0.656 / 21.43                                 \\
		& $\spadesuit$IMDer*                    & 0.689 / 21.80                                 & 0.589 / 28.30                                 & 0.510 / 30.28                                 & 0.576 / 30.45                                 & 0.549 / \textbf{23.06}                                  & 0.345 / 41.30                                 & 0.551 / 29.20                                 & 0.632 / 23.04                                 \\
		& $\spadesuit$MoMKE*                  & 0.680 / 19.88                                 & 0.620 / \textbf{26.70}                                 & 0.554 / \textbf{25.58}                                 & 0.600 / 32.74                                 & 0.555 / 33.12                                  & 0.455 / 38.65                                 & 0.581 / 29.45                                 & 0.676 / 21.87                                 \\
		& $\spadesuit$SDR-GNN*                   & 0.678 / \textbf{19.01}                                 & 0.618 / 27.21                                 & 0.569 / 26.26                                 & \textbf{0.619 / 26.41}                                 & 0.579 / 30.84                                  & 0.417 / 40.16                                 & \textbf{0.585} / 28.31                                 & 0.672 / 20.83                                 \\
		& \textbf{MCMoE(Ours)}     & {\color{red} \textbf{0.719 / 16.54}} & {\color{red} \textbf{0.708 / 21.51}} & {\color{red} \textbf{0.677 / 23.38}} & {\color{red} \textbf{0.677 / 21.09}} & \textbf{0.633} / {\color{red} \textbf{19.99}}                         & {\color{red} \textbf{0.508}} / \textbf{38.24} & {\color{red} \textbf{0.659 / 23.46}} & {\color{red} \textbf{0.759 / 18.50}} \\
		\multirow{-9}{*}{TES}        & ${\Delta _{SOTA}}$                      & {\color{blue} ↑0.3\% / ↓13.0\%}       & {\color{blue} ↑8.6\% / ↓19.4\%}       & {\color{blue} ↑1.7\% / ↓8.6\%}        & {\color{blue} ↑9.4\% / ↓20.1\%}      &  {\color{gray} ↓2.2\%} / {\color{blue}↓13.3\%}        & {\color{blue} ↑11.4\%} / {\color{gray} ↑3.0\%}       & {\color{blue} ↑12.6\% / ↓17.1\%}      & {\color{blue} ↑0.7\% / ↓7.8\%}        \\ \midrule
		& $\heartsuit$MLP-Mixer*                 & 0.789 / \textbf{8.31}                                  & 0.736 / \textbf{11.67}                                 & 0.608 / 15.43                                 & 0.747 / \textbf{10.66}                                 & 0.614 / 13.97                                  & 0.486 / 20.67                                 & 0.676 / \textbf{13.45}                                 & 0.824 / \textbf{7.88}                                  \\
		& $\heartsuit$PAMFN                     & \textbf{0.862} / 8.45                                  & \textbf{0.811} / 28.71                                 & 0.574 / 51.63                                 & \textbf{0.810} / 29.14                                 & 0.584 / 58.51                                  & 0.079 / 83.12                                 & 0.679 / 43.26                                 &\textbf{ 0.872} / 8.16                                  \\
		& $\clubsuit$ActionMAE*                 & 0.768 / 23.84                                 & 0.702 / 15.32                                 & 0.655 / 19.04                                 & 0.641 / 18.75                                 & 0.570 / 19.22                                  & 0.514 / 21.44                                 & 0.649 / 19.60                                 & 0.745 / 11.07                                 \\
		& $\spadesuit$GCNet*                     & 0.788 / 20.05                                 & 0.705 / 14.61                                 & 0.632 / 16.28                                 & 0.724 / 14.54                                 & 0.584 / 14.71                                  & 0.516 / 26.37                                 & 0.668 / 17.76                                 & 0.735 / 12.43                                 \\
		& $\spadesuit$IMDer*                     & 0.797 / 8.58                                  & 0.718 / 17.01                                 & 0.621 / 17.70                                 & 0.755 / 20.44                                 & 0.679 / 29.71                                  & 0.462 / 22.63                                 & 0.685 / 19.34                                 & 0.762 / 11.00                                 \\
		& $\spadesuit$MoMKE*                  & 0.813 / 9.81                                  & 0.747 / 14.50                                 & \textbf{0.723 / 13.66 }                                & 0.753 / 13.57                                 &\textbf{ 0.734 / 11.81}                                  & \textbf{0.537 / 19.54}                                 & \textbf{0.727} / 13.81                                 & 0.805 / 12.74                                 \\
		& $\spadesuit$SDR-GNN*                   & 0.811 / 10.97                                 & 0.733 / 14.04                                 & 0.664 / 15.51                                 & 0.748 / 14.11                                 & 0.707 / 12.15                                  & 0.536 / 24.11                                 & 0.709 / 15.15                                 & 0.784 / 12.06                                 \\
		& \textbf{MCMoE(Ours)}     & {\color{red} \textbf{0.878 / 5.49}}  & {\color{red} \textbf{0.847 / 7.76}}  & {\color{red} \textbf{0.770 / 11.44}} & {\color{red} \textbf{0.831 / 9.19}}  & {\color{red} \textbf{0.753 / 10.79}}  & {\color{red} \textbf{0.603 / 18.84}} & {\color{red} \textbf{0.795 / 10.58}} & {\color{red} \textbf{0.880 / 5.81}}  \\
		\multirow{-9}{*}{PCS}        & ${\Delta _{SOTA}}$                      & {\color{blue} ↑1.9\% / ↓33.9\%}       & {\color{blue} ↑4.4\% / ↓33.5\%}       & {\color{blue} ↑6.5\% / ↓16.3\%}       & {\color{blue} ↑2.6\% / ↓13.8\%}       & {\color{blue} ↑2.6\% / ↓8.6\%}         & {\color{blue} ↑12.3\% / ↓3.6\%}       & {\color{blue} ↑9.4\% / ↓21.3\%}       & {\color{blue} ↑0.9\% / ↓26.3\%}       \\ \bottomrule
	\end{tabular}}
	\vspace{-8pt}
    \caption{omparisons with the state-of-the-art on the Fis-V under incomplete multimodal scenarios. $v$, $f$, and $a$ refer to the RGB, flow, and audio modalities. ``Average'' denotes the average result of all six incomplete multimodal combinations. The {\color{red} \textbf{red bold}} / \textbf{black bold} indicate the best / second-best results. * indicates our reimplementation. ${\Delta _{SOTA}}$ means the performance {\color{blue} increase} or {\color{gray} decrease} of our MCMoE compared to the best competing methods. $\heartsuit$, $\clubsuit$, and $\spadesuit$ mean the evaluated method sources for multimodal AQA, incomplete multimodal action recognition, and incomplete multimodal emotion recognition.}
	\vspace{-1pt}
    \label{tab:a4}
\end{table*}

\begin{table*}[t]
	\centering
	\tabcolsep=1.5pt
	\resizebox{\linewidth}{!}{
	\begin{tabular}{@{}c|c|cccccccc@{}}
		\toprule
		&        & \multicolumn{8}{c}{Testing Condition (Spearman Correlation (↑) / Mean Square Error (↓))}      \\ \cmidrule(l{0pt}r{0pt}){3-10}
		\multirow{-2}{*}{Types} & \multirow{-2}{*}{Methods}& $\{v,f\}$          & $\{v,a\}$           & $\{f,a\}$           & $\{v\}$             & $\{f\}$            & $\{a\}$             & Average           & $\{v,f,a\}$                              \\ \midrule
		& $\heartsuit$MLP-Mixer*                 & 0.662 / 9.52                                 & 0.618 / 9.78                                 & 0.464 / 11.33                                & 0.561 / 13.76                                & 0.439 / 14.81                                  & 0.359 / 18.33                                 & 0.525 / 12.92                                & 0.694 / 8.09                                 \\
		& $\heartsuit$PAMFN                     &\textbf{ 0.739 / 6.90}                                 & \textbf{0.702 / 8.24 }                                & 0.236 / 149.42                               & 0.624 / 8.77                                 & 0.343 / 144.22 & 0.157 / 144.63                                & 0.501 / 77.03                                & \textbf{0.757 / 6.24}                                 \\
		& $\clubsuit$ActionMAE*                & 0.650 / 8.13                                 & 0.554 / 10.30                                & 0.530 / 10.50                                & 0.615 / 9.91                                 & 0.471 / 9.45                                   & 0.367 / 12.37                                 & 0.537 / 10.11                                & 0.613 / 8.77                                 \\
		& $\spadesuit$GCNet*                    & 0.627 / 8.02                                 & 0.621 / 7.70                                 & 0.485 / 9.21                                 & 0.598 / \textbf{8.39}                                 & 0.446 / 10.37                                  & \textbf{0.422} / 15.44                                 & 0.539 / 9.85                                 & 0.616 / 7.31                                 \\
		& $\spadesuit$IMDer*                     & 0.646 / 7.80                                 & 0.596 / 8.27                                 & 0.527 / 8.36                                 & 0.623 / 9.10                                 & 0.532 / 8.53                                   & 0.333 / 12.33                                 & 0.550 / 9.06                                & 0.654 / 7.30                                 \\
		& $\spadesuit$MoMKE*                   & 0.652 / 7.74                                 & 0.592 / 9.09                                 & 0.603 / \textbf{7.30}                                 & 0.524 / 10.75                                & \textbf{0.564 / 7.83}                                   & {\color{red} \textbf{0.442}} /\textbf{ 11.75}                                 & 0.566 / 9.08                                 & 0.670 / 7.47                                 \\
		& $\spadesuit$SDR-GNN*                   & 0.646 / 6.94                                 & 0.611 / 8.49                                 & \textbf{0.624} / 8.11                                 & \textbf{0.631} / 9.01                                 & 0.473 / 8.48                                  & 0.439 / 12.08                                 & \textbf{0.576 / 8.85}                                 & 0.643 / 7.64                                 \\
		& \textbf{MCMoE(Ours)}     & {\color{red} \textbf{0.803 / 6.05}} & {\color{red} \textbf{0.732 / 7.45}} & {\color{red} \textbf{0.643 / 7.96}} & {\color{red} \textbf{0.719 / 8.28}} & {\color{red} \textbf{0.639 / 8.22}}   & 0.327 / {\color{red} \textbf{11.70}} & {\color{red} \textbf{0.664 / 8.28}} & {\color{red} \textbf{0.806 / 5.66}} \\
		\multirow{-9}{*}{Ball}       & ${\Delta _{SOTA}}$                      & {\color{blue} ↑8.7\% / ↓12.3\%}      & {\color{blue} ↑4.3\% / ↓3.2\%}       & {\color{blue} ↑3.0\%} / {\color{gray} ↑9.0\%}       & {\color{blue} ↑13.9\% / ↓1.3\%}      & {\color{blue} ↑13.3\% / ↑5.0\%}        & {\color{gray} ↓26.0\%} / {\color{blue} ↓0.4\%}       & {\color{blue} ↑15.3\% / ↓6.4\%}      & {\color{blue} ↑6.5\% / ↓9.3\%}       \\\midrule
		& $\heartsuit$MLP-Mixer*                 & 0.768 / 4.96                                 & 0.468 / 9.86                                 & 0.331 / 10.91                                & 0.714 / 5.96                                 & 0.599 / 7.15                                   & \textbf{0.131} / 17.82                                 & 0.535 / 9.44                                 & 0.736 / 6.13                                 \\
		& $\heartsuit$PAMFN                     & 0.747 / 5.36                                 & 0.582 / 8.48                                 & 0.362 / 115.32                               & 0.703 / 8.01                                 & 0.519 / 127.63                                 & {\color{red} \textbf{0.320}} / 118.72                                & 0.559 / 63.92                                & {\color{red} \textbf{0.825}} / 7.45                                 \\
		& $\clubsuit$ActionMAE*                 & 0.778 / 5.25                                 & 0.590 / 8.36                                 & 0.433 / \textbf{9.43}                                 & 0.734 / 16.29                                & 0.436 / 14.37                                  & 0.117 / {\color{red} \textbf{8.49}}                                  & 0.549 / 10.37                                & 0.668 / 6.35                                 \\
		& $\spadesuit$GCNet*                     & 0.781 / 5.51                                 & 0.628 / 7.36                                 & 0.477 / 10.43                                & 0.729 / 6.36                                 & 0.622 / 8.68                                   & 0.026 / \textbf{11.96}                                 & 0.581 / 8.38                                 & 0.709 / 7.11                                 \\
		& $\spadesuit$IMDer*                    & 0.792 / 5.43                                 & 0.615 / 7.61                                 & 0.405 / 9.66                                 & 0.729 / \textbf{5.43}                                 & 0.634 / \textbf{8.05}                                   & 0.069 / 12.34                                 & 0.579 / \textbf{8.09}                                 & 0.698 / \textbf{5.80}                                  \\
		& $\spadesuit$MoMKE*                  & 0.783 / 4.65                                & 0.589 / 8.14                                 & \textbf{0.509} / 11.23                                & 0.731 / 7.12                                 & \textbf{0.654} / 10.05                                  & 0.056 / 12.31                                 & 0.589 / 8.92                                 & 0.709 / 5.91                                 \\
		& $\spadesuit$SDR-GNN*                   & \textbf{0.797 / 4.47}                                 & \textbf{0.647 / 7.14}                                 & 0.492 / 10.13                                & \textbf{0.751} / 6.07                                 & 0.641 / 8.34                                  & 0.047 / 12.62                                 & \textbf{0.602} / 8.13                                 & 0.730 / 6.60                                 \\
		& \textbf{MCMoE(Ours)}     & {\color{red} \textbf{0.820 / 4.10}} & {\color{red} \textbf{0.720 / 5.58}} & {\color{red} \textbf{0.633 / 7.22}} & {\color{red} \textbf{0.773 / 4.85}} & {\color{red} \textbf{0.657 / 6.82}}   & 0.005 / 14.25 & {\color{red} \textbf{0.648 / 7.14}} & \textbf{0.815} / {\color{red} \textbf{4.22}} \\
		\multirow{-9}{*}{Clubs}      & ${\Delta _{SOTA}}$                      & {\color{blue} ↑2.9\% / ↓8.3\%}      & {\color{blue} ↑11.3\% / ↓21.8\%}     & {\color{blue} ↑24.4\% / ↓23.4\%}     & {\color{blue} ↑2.9\% / ↓10.7\%}      & {\color{blue} ↑0.5\% / ↓4.6\%}         & {\color{gray} ↓98.4\% / ↑67.8\%}      & {\color{blue} ↑7.6\% / ↓11.7\%}     & {\color{gray} ↓1.2\%} / {\color{blue}↓27.2\%}      \\ \midrule
		& $\heartsuit$MLP-Mixer*                & 0.701 / 8.08                                 & 0.682 / 27.60                                & 0.468 / \textbf{9.84}                                 & 0.641 / 7.35                                 & 0.525 / 13.97                                  & 0.122 / \textbf{13.29}                                 & 0.546 / 13.35                                & 0.787 / 7.68                                 \\
		& $\heartsuit$PAMFN                    & 0.767 / 5.42                                 & 0.733 / 10.08                                & 0.524 / 201.27                               & 0.702 / 7.53                                 &\textbf{ 0.600} / 202.95                                 & 0.042 / 204.06                                & 0.598 / 105.22                               & \textbf{0.836} / {\color{red} \textbf{5.21}  }                               \\
		& $\clubsuit$ActionMAE*                & 0.737 / 5.64                                 & 0.692 / 7.02                                 & 0.595 / 12.44                                & 0.736 / \textbf{7.28}                                 & 0.504 / \textbf{11.53}                                  & 0.202 / 33.18                                 & 0.602 / 12.85                                & 0.771 / 6.25                                 \\
		& $\spadesuit$GCNet*                     & 0.734 / 6.75                                 & 0.646 / 7.61                                 & 0.549 / 19.14                                & 0.756 / 10.16                                & 0.520 / 115.81                                 & 0.222 / 17.82                                 & 0.595 / 29.55                                & 0.732 / 5.93                                 \\
		& $\spadesuit$IMDer*                     & 0.769 / \textbf{4.31}                                 & 0.690 / 6.99                                 & 0.517 / \textbf{9.84}                                 & 0.761 / 8.38                                 & 0.531 / 11.93                                  & 0.333 / 18.98                                 & 0.622 / 10.07                                & 0.754 / 6.41                                 \\
		& $\spadesuit$MoMKE*                   &\textbf{ 0.780} / 4.93                                 & \textbf{0.756 / 6.60  }                               &\textbf{ 0.602} / 10.12                                & {\color{red} \textbf{0.779}} / 7.34                                 & 0.552 / 12.83                                  & \textbf{0.354} / 13.36                                 & \textbf{0.661 / 9.20 }                                & 0.789 / 6.01                                 \\
		& $\spadesuit$SDR-GNN*                   & 0.765 / 6.85                                 & 0.681 / 7.02                                 & 0.572 / 11.44                                & \textbf{0.778} / 8.68                                 & 0.541 / 13.49                                 & 0.331 / 14.34                                 & 0.633 / 10.30                                & 0.763 / 5.76                                 \\
		& \textbf{MCMoE(Ours)}     & {\color{red} \textbf{0.796 / 6.59}} & {\color{red} \textbf{0.809 / 5.66}} & {\color{red} \textbf{0.771 / 8.62}} & 0.757 / {\color{red} \textbf{6.53}}                                 & {\color{red} \textbf{0.643 / 10.53}}  & {\color{red} \textbf{0.490 / 12.06}} & {\color{red} \textbf{0.726 / 8.33}} & {\color{red} \textbf{0.845}} / \textbf{5.62} \\
		\multirow{-9}{*}{Hoop}       & ${\Delta _{SOTA}}$                      & {\color{blue} ↑2.1\% / ↑52.9\%}      & {\color{blue} ↑7.0\% / ↓14.2\%}      & {\color{blue} ↑28.1\% / ↓12.4\%}     & {\color{gray} ↓2.8\%} / {\color{blue}↓10.3\%}      & {\color{blue} ↑7.2\% / ↓8.7\%}         & {\color{blue} ↑38.4\% / ↓9.3\%}       & {\color{blue} ↑9.8\% / ↓9.5\%}       & {\color{blue} ↑1.1\%} / {\color{gray}↑7.9\%}       \\\midrule
		& $\heartsuit$MLP-Mixer*                 & 0.787 / 6.37                                 & 0.663 / 9.11                                 & 0.644 / 8.65                                 & 0.690 / 11.62                                & 0.676 / 9.88                                   & {\color{red}\textbf{0.350}} / 14.61                                 & 0.651 / 10.04                                & 0.789 / 8.03                                 \\
		& $\heartsuit$PAMFN                     & 0.798 / 9.42                                 & 0.381 / 128.67                               & 0.622 / 22.44                                & 0.594 / 135.13                               & 0.449 / 18.95                                  & -0.004 / 139.33                               & 0.511 / 75.66                                & \textbf{0.846} / 7.67                                 \\
		& $\clubsuit$ActionMAE*                 & 0.719 / 10.17                                & 0.635 / 9.37                                 & 0.609 / 9.18                                 & 0.658 / 16.15                                & 0.651 / 9.79                                   & \textbf{0.309 / 13.31}                                 & 0.609 / 11.33                                & 0.759 / 6.69                                 \\
		& $\spadesuit$GCNet*                    & 0.785 / 6.46                                 & 0.656 / 9.13                                 & 0.688 / 8.22                                 & 0.702 / 7.55                                 & 0.660 / 9.18                                   & 0.210 / 15.57                                 & 0.642 / 9.35                                 & 0.786 / 5.46                                 \\
		& $\spadesuit$IMDer*                     & 0.757 / 6.57                                 & \textbf{0.676} / 7.34                                 & \textbf{0.759} / {\color{red} \textbf{7.33}}                                 & 0.667 / 7.45                                & 0.673 / \textbf{8.88}                                   & 0.073 / {\color{red} \textbf{13.12}}                                 & 0.635 / \textbf{8.45}                                 & 0.776 / 5.96                                 \\
		& $\spadesuit$MoMKE*                   & \textbf{0.809 / 5.42}                                 & 0.662 / 8.07                                 & {\color{red} \textbf{0.764}} / 8.92                                 & 0.695 / 8.48                                 & \textbf{0.699} / 9.60                                   & 0.180 / 15.57                                 & 0.667 / 9.34                                 & 0.800 / \textbf{5.34 }                                \\
		& $\spadesuit$SDR-GNN*                   & 0.801 / 6.07                                 & \textbf{0.676 / 6.86}                                 & 0.729 / \textbf{7.91}                                 & \textbf{0.730 / 7.33}                                 & 0.686 / 8.91                                  & 0.217 / 15.07                                 & \textbf{0.668} / 8.69                                 & 0.810 / 5.41                                 \\
		& \textbf{MCMoE(Ours)}     & {\color{red} \textbf{0.861 / 4.57}} & {\color{red} \textbf{0.841 / 4.64}} & 0.730 / 8.80                                 & {\color{red} \textbf{0.811 / 5.33}} & {\color{red} \textbf{0.706 / 8.79}}   & 0.256 / 14.78                                 & {\color{red} \textbf{0.741 / 7.82}} & {\color{red} \textbf{0.890 / 3.89}} \\
		\multirow{-9}{*}{Ribbon}     & ${\Delta _{SOTA}}$                      & {\color{blue} ↑6.4\% / ↓15.7\%}      & {\color{blue} ↑24.4\% / ↓32.4\%}     & {\color{gray} ↓4.5\% / ↑20.1\%}      & {\color{blue} ↑11.1\% / ↓27.3\%}     & {\color{blue} ↑1.0\% / ↓1.0\%}         & {\color{gray} ↓26.9\% / ↑12.7\%}      & {\color{blue} ↑10.9\% / ↓7.5\%}      & {\color{blue} ↑5.2\% / ↓27.2\%}     \\ \bottomrule
	\end{tabular}}
	\vspace{-8pt}
    \caption{omparisons with the state-of-the-art on the RG under incomplete multimodal scenarios. $v$, $f$, and $a$ refer to the RGB, flow, and audio modalities. ``Average'' denotes the average result of all six incomplete multimodal combinations. The {\color{red} \textbf{red bold}} / \textbf{black bold} indicate the best / second-best results. * indicates our reimplementation. ${\Delta _{SOTA}}$ means the performance {\color{blue} increase} or {\color{gray} decrease} of our MCMoE compared to the best competing methods. $\heartsuit$, $\clubsuit$, and $\spadesuit$ mean the evaluated method sources for multimodal AQA, incomplete multimodal action recognition, and incomplete multimodal emotion recognition.}
	\vspace{-8pt}
    \label{tab:a5}
\end{table*}

\bibliography{aaai2026}

\end{document}